%% file: main.tex
\documentclass[conference]{IEEEtran}
\usepackage{cite}

\ifCLASSINFOpdf
\else
\fi

\newcommand{\sys}{{\sc URVFL}\xspace}
\usepackage{derivative}
\usepackage{mathptmx}
\usepackage[ruled,lined]{algorithm2e}
\usepackage{microtype}
\usepackage{graphicx}
\usepackage{booktabs} 
\usepackage{hyperref}
\usepackage{stfloats}
\usepackage{bbm}
\usepackage{amsthm}
\usepackage{multirow}
\usepackage{tabularx}
\usepackage{amsmath}
\usepackage{diagbox} 
\usepackage{booktabs} 
\usepackage{siunitx}  
\usepackage{diagbox} 
\usepackage{enumitem}
\usepackage{subcaption}
\usepackage{amssymb, amsmath, amsthm}
\usepackage[mathcal]{eucal}
\usepackage{chngcntr}
\usepackage{dsfont}
\usepackage{verbatim}
\usepackage{color}
\usepackage{xspace}
\SetKwInput{KwInput}{Data}                
\SetKwInput{KwOutput}{Initialization}  
\usepackage{adjustbox}
\usepackage{placeins}

\usepackage{float}  
\usepackage{hyperref}
\hypersetup{
    colorlinks=true,
    linkcolor=blue,    
    citecolor=blue,    
    urlcolor=blue      
}

\begin{document}
%
\title{URVFL: Undetectable Data Reconstruction Attack on Vertical Federated Learning}

\author{
    Duanyi Yao$^*$, Songze Li$^{\dagger\ddagger}$, Xueluan Gong$^{\S}$, Sizai Hou$^*$, Gaoning Pan$^{\P}$\\[2ex]
    $^*$ The Hong Kong University of Science and Technology\\[1ex]
    $^{\dagger}$ Southeast University\\[1ex]
    $^{\ddagger}$ Engineering Research Center of Blockchain Application, Supervision and Management (Southeast University),\\
    Ministry of Education\\[1ex]
    $^{\S}$ Wuhan University\\[1ex]
    $^{\P}$ Hangzhou Dianzi University
}


%



\IEEEoverridecommandlockouts
\makeatletter\def\@IEEEpubidpullup{6.5\baselineskip}\makeatother
\IEEEpubid{\parbox{\columnwidth}{
		Network and Distributedx System Security (NDSS) Symposium 2025\\
		 24 - 28 February 2025, San Diego, CA, USA\\
		ISBN 979-8-9894372-8-3\\
		https://dx.doi.org/10.14722/ndss.2025.24046\\
		www.ndss-symposium.org
}
\hspace{\columnsep}\makebox[\columnwidth]{}}

\maketitle

\begin{abstract}
  Vertical Federated Learning (VFL) is a collaborative learning paradigm designed for scenarios where multiple clients share disjoint features of the same set of data samples. Albeit a wide range of applications, VFL is faced with privacy leakage from data reconstruction attacks.   
  These attacks generally fall into two categories: honest-but-curious (HBC), where adversaries steal data while adhering to the protocol; and malicious attacks, where adversaries breach the training protocol for significant data leakage. While most research has focused on HBC scenarios, the exploration of malicious attacks remains limited. 
  
Launching effective malicious attacks in VFL presents unique challenges: 1) Firstly, given the distributed nature of clients' data features and models, each client rigorously guards its privacy and prohibits direct querying, complicating any attempts to steal data;
2) Existing malicious attacks alter the underlying VFL training task, and are hence easily detected by comparing the received gradients with the ones received in honest training. 
To overcome these challenges, we develop \sys, a novel attack strategy that evades current detection mechanisms. The key idea is to integrate a \textit{discriminator with auxiliary classifier} that takes a full advantage of the label information and generates malicious gradients to the victim clients: on one hand, label information helps to better characterize embeddings of samples from distinct classes, yielding an improved reconstruction performance; on the other hand, computing malicious gradients with label information better mimics the honest training, making the malicious gradients indistinguishable from the honest ones, and the attack much more stealthy.
Our comprehensive experiments demonstrate that \sys significantly outperforms existing attacks, and successfully circumvents SOTA detection methods for malicious attacks. Additional ablation studies and evaluations on defenses further underscore the robustness and effectiveness of \sys. Our code will be available at  \href{URVFL}{https://github.com/duanyiyao/URVFL}{}.


\end{abstract}

\input{introduction}
\input{background}
\input{Preliminary}
\input{challenge}

\input{methodology}
\input{experiment}

\input{ablationstudy}
\input{conclusion}



\section*{Acknowledgment}
    The work of Songze Li is in part supported by the National Nature Science Foundation of China (NSFC) Grant 62106057, and the Fundamental Research Funds for the Central Universities (Grant No. 2242024k30059).

\bibliographystyle{IEEEtranS}
\bibliography{references.bib}
\newpage
\appendices
\input{appendix}\label{appendix}

\end{document}

%% file: introduction.tex
\section{Intoduction}

Federated learning (FL) is an emerging privacy-preserving collaborative learning paradigm, which allows multiple distributed clients to securely train a model without sharing private data~\cite{yang2019federated}. FL can be categorized into horizontal FL (HFL) and vertical FL (VFL)~\cite{liu2024vertical}. In HFL, different clients possess distinct sets of training samples, yet all samples share the same feature space. In contrast, VFL is applicable to scenarios where clients have the same sample identifiers but possess disjoint feature spaces, e.g., an finance company may hold investment records for a set of customers, while a bank may have the expenditure details for the same set of customers. Due to its capability to integrate diverse data sources, VFL is increasingly gaining attention and has been applied in various fields, including finance~\cite{chen2020vafl}, healthcare~\cite{poirot2019split}, and recommendation systems~\cite{yang2020federated}.  

\begin{figure}[htbp]
    \centering
    \includegraphics[width=0.9\linewidth]{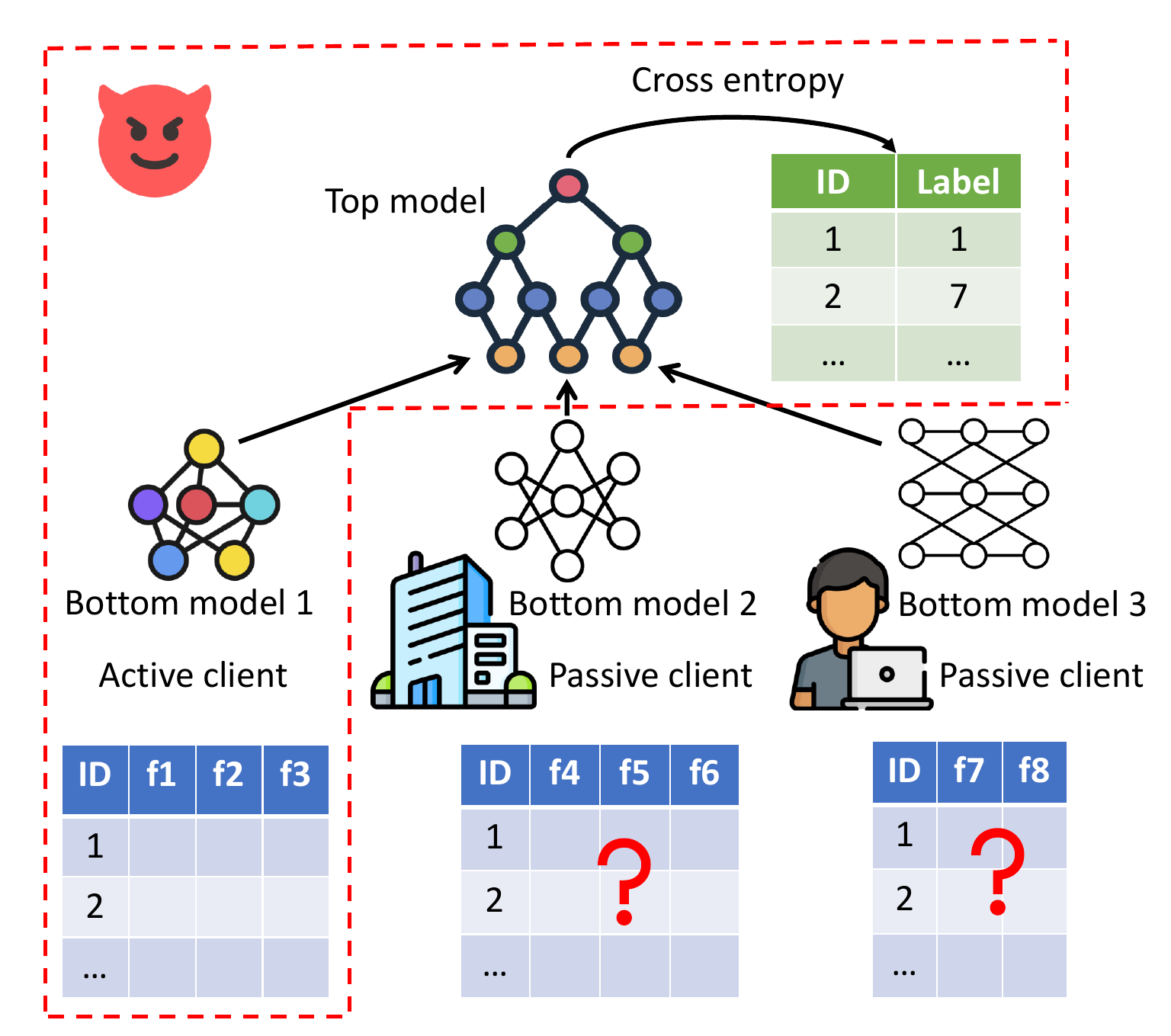}
    \caption{Illustration of a VFL system with one active client and two passive clients. The area enclosed by the red dashed lines contains the information accessible to and the actions taken by the active client. In a data reconstruction attack, the malicious active client intends to recover the private features of the target passive clients.}
    \label{fig:vfl}
\end{figure}

In VFL~\cite{liu2022vertical, chen2020vafl, li2023fedvs, thapa2022splitfed, feng2020multi, wu2020privacy, yang2019quasi}, an active client holds both labels and partial features, while multiple passive clients hold disjoint features for the same samples. VFL typically employs a split learning framework~\cite{vepakomma2018split, abuadbba2020can, thapa2022splitfed, kim2020multiple}, and utilizes separate models to process these disjoint features. Specifically, a split VFL system comprises one top model and several bottom models: each passive client operates a bottom model, whereas the active client manages both a bottom model and a top model (see Figure~\ref{fig:vfl}). During the training phase, for each batch of training samples, all passive and active clients compute embeddings using their respective bottom models; these embeddings are then transmitted to the active client, who outputs the predictions with the top model and computes the loss from the labels. Following this, the gradients for embeddings are sent back to all clients to update their bottom models.

\noindent\textbf{Data reconstruction attacks on VFL:} While VFL emphasizes privacy by sharing embeddings instead of raw data, emerging data reconstruction attacks have revealed potential vulnerabilities that could lead to privacy leakage within VFL models. Such attacks are categorized into honest-but-curious (HBC) and malicious attacks. HBC adversaries stealthily extract private features while adhering to the training protocol, whereas malicious adversaries actively manipulate or violate the protocol to steal private features. 
Notable methods like GRNA~\cite{luo2021feature} and GIA~\cite{jiang2022comprehensive} focus on HBC scenarios. In GRNA, under the assumption that the adversary can query the passive clients' models, target features could be inferred by training a generative regression network to learn correlations between target features and the adversary's partial features, thus enabling generation of target features. GIA~\cite{jiang2022comprehensive}, presupposing the adversary can inject adversarial features to the target clients, employs optimization techniques to reconstruct the target features.  A malicious data reconstruction attack in VFL, called AGN, is developed in \cite{vu2023active}, where the adversary transmits malicious gradients to passive clients to induce more significant feature leakage. 

A class of closely related attacks have also been developed for split learning (SL), which can be viewed as a special case of VFL.
In SL, there is a clear division of roles where an active client, holding a top model and labels, but without partial features, collaborates with a single passive client who manages a bottom model and all data samples with complete features. In HBC scenario, attacks like PCAT~\cite{gao2023pcat} and SDAR~\cite{zhu2023passive} use the top model to train a shadow model on an auxiliary dataset, which is subsequently used to train a decoder for data reconstruction. In the malicious scenario, FSHA~\cite{pasquini2021unleashing} manipulates the training objective, generating malicious gradients to achieve more accurate data reconstruction.

\noindent\textbf{Challenges of existing data reconstruction attacks on VFL:} While malicious attacks could be more detrimental to privacy leakage in VFL, research works on malicious attacks remains limited. We identify two primary challenges to launch effective malicious attacks: 1) \textit{Distributed features and limited model access.} The heterogeneous nature of feature sources across distributed clients, together with the lack of access to passive clients' local models, 
makes it difficult for the adversary to infer target features using its own partial features; 2) \textit{Powerful detections.} Existing malicious attacks, such as FSHA and AGN, 
replace the honest training task with a malicious task, causing the gradients returned to passive clients to appear different from those in honest training, which renders these attacks easily detectable. SOTA detection strategies for malicious attacks, e.g., SplitGuard~\cite{erdogan2022splitguard} and Gradient Scrutinizer~\cite{fufocusing}, monitor incoming gradients to identify potential attacks and significantly diminish the success of malicious data reconstruction attacks.

\noindent\textbf{Our attack URVFL:} In response to these challenges, we investigate whether a malicious attacker in VFL can circumvent detection measures to maintain effective data reconstruction. We answer this question affirmatively, by developing a novel undetectable data reconstruction attack on VFL (referred to as \sys).
\sys begins with the adversary pretraining an encoder, a decoder, and its bottom model to achieve high reconstruction performance on an auxiliary data. 
Following this, the adversary freezes the encoder and bottom model, and trains a discriminator with auxiliary classifier (DAC)~\cite{hou2022conditional} that integrates labels information to generate malicious gradients transmitted to passive clients. These malicious gradients can transfer embeddings distribution from the encoder to passive clients' models. In data reconstruction phase, the trained decoder is leveraged to reconstruct target features from the target clients, as these passive clients' models are meticulously guided to mimic the encoder by DAC.

To address challenge 1, \sys captures correlations between the adversary's partial features and the target features without direct queries or data injections, using DAC to generate malicious gradients that indirectly guide the passive clients' model to leak feature information. To handle challenge 2, we use DAC to incorporate label information into the computation of the malicious loss function, generating 
malicious gradients almost \textit{indistinguishable} from the gradients received in honest training, which helps \sys to effectively circumvent current detection strategies. Additionally, these label information helps to better transfer the embedding distribution from the encoder to the passive clients' models compared with a traditional discriminator, and provide extra information for achieving a smaller reconstruction error.
 As a result, over extensive empirical evaluations, \sys is shown to achieve better reconstruction performance than SOTA attacks, and successfully circumvents all existing detections against malicious attacks. Further ablation studies and evaluations against defense strategies underscore the robustness and effectiveness of \sys.

In summary, this paper makes the following main contributions:
\begin{enumerate}[leftmargin=*]
\item We introduce \sys, a novel undetectable data reconstruction attack on VFL, designed to bypass existing detection measures against malicious attacks and effectively steal private features.
\item We develop a DAC as part of \sys, which:
\begin{itemize}
\item Significantly enhances the effectiveness of embedding distribution transfer and overall attack performance.
\item Generates malicious gradients that are indistinguishable from those produced during honest training.
\end{itemize}
\item We conduct rigorous evaluations of \sys using five representative datasets, demonstrating that it:
\begin{itemize}
\item Circumvents SOTA detection strategies against malicious data reconstruction.
\item Outperforms SOTA data reconstruction attacks on VFL, both with and without detection mechanisms in place.
\item Achieves high-quality reconstruction despite defense methods designed for both malicious and HBC attacks in VFL.
\end{itemize}
\end{enumerate}

%% file: background.tex
\section{Background and Related works}
In this section, we review SOTA data reconstruction attacks on VFL, and existing detection and defense mechanisms for these attacks. 


\subsection{Data reconstruction attacks on VFL}
Although VFL safeguards feature privacy by exchanging embeddings and intermediate gradients instead of raw features, recent research has revealed its vulnerability to data reconstruction attacks that can lead to private feature leakage.  
 In HBC setting, the attacks described in~\cite{luo2021feature} and~\cite{yang2023practical} assume that an HBC adversary can query the passive clients' bottom model. 
During the inference phase, the adversary covertly trains a generative model that is fed random vectors and partial data features available to the adversary, thereby producing synthetic inputs. By minimizing the discrepancy of the bottom models' output between synthetic inputs and target features, the generative model is trained to approximate these features accurately. 
While the adversary in~\cite{luo2021feature} can access the gradients of the target bottom model directly, the adversary in~\cite{yang2023practical} employs zeroth-order optimization to estimate these gradients, thereby bypassing the need for direct gradient access.

The approach in~\cite{jiang2022comprehensive}, GIA, operates under a black-box scenario where the adversarial active client lacks direct access to the passive clients' bottom model. Despite this limitation, the adversary can strategically insert a batch of auxiliary data features into the datasets of the passive clients and subsequently collect the resulting output embeddings. The adversary then trains a shadow model using the collected auxiliary features and embeddings pairs. After training, the adversary optimizes noise to closely approximate the target features by querying this shadow model.
It is essential to emphasize that these attack methodologies, while effective in reconstructing private data, depend on the adversary having either direct access to the passive clients' models or the capability to inject data into these clients.

In malicious setting, the attack method, AGN~\cite{vu2023active}, utilizes a generator to create fake target features from passive clients' embeddings. These fake features are then evaluated by a discriminator, which distinguishes between real and fake features to generate malicious gradients. These gradients are sent back to the passive clients, aiming to train embeddings that facilitate the reconstruction of target features. Despite the innovative approach of AGN, our evaluations in the section of experiments reveal that the data reconstructed by this method is relatively coarse and remains susceptible to existing defense mechanisms.

\noindent \textbf{Data reconstruction attacks on split learning.} A parallel strand of privacy attacks on SL is adaptable to VFL with certain modifications. Unlike VFL which involves multiple clients, SL typically includes only one passive client and an active client managing the bottom and top models, respectively, which can be viewed as a special case of VFL. 

For HBC settings, He et al. pioneer privacy attacks on SL in \cite{he2019model}, exploring various levels of adversarial knowledge. In a black-box scenario, where the adversary lacks the knowledge of the bottom model but can make queries, an inverse model is used to reconstruct private features. In scenarios without query capabilities, the adversary utilizes a shadow model trained with auxiliary data under the guidance of the top model. Similar with GIA, the shadow model is then used to reconstruct target features through optimization. Unsplit~\cite{erdougan2022unsplit} presents a scenario where the adversary knows the bottom model's structure but lacks auxiliary data. This method involves optimizing synthetic features and the shadow model concurrently to reduce disparities between real and synthetic features' embeddings. PCAT~\cite{gao2023pcat} depicts an adversary with access to a batch of auxiliary data but unaware of the bottom model's structure and parameters. The strategy involves covertly training a shadow model with the auxiliary data, guided by the top model in each training round. Then an inverse model is trained on the shadow model for data reconstruction.
SDAR~\cite{zhu2023passive}, assuming knowledge of the bottom model's architecture, utilizes auxiliary data and adversarial regularization to train both a shadow model and its inverse model. SDAR employs two discriminators for adversarial regularization: one differentiates between the embeddings of synthetic and genuine features, while the other identifies the data's origin, whether it comes from the inverse model or the auxiliary data.
For malicious setting, the attack method, FSHA~\cite{pasquini2021unleashing}, replaces the top model with a discriminator for aligning the bottom model with a local encoder. Concurrently, an decoder trained on the encoder can apply to reconstruct target features.

\subsection{Defenses and detections against data reconstruction attacks}
\noindent\textbf{Defenses:} As data leakage risks in VFL escalate, a variety of defensive strategies have been developed to defend data reconstruction attacks. These defenses, particularly against both HBC and malicious adversary, can be broadly classified into cryptographic methods, perturbation-based methods, and adversarial training methods. Cryptographic strategies include homomorphic encryption (HE)\cite{xu2021efficient, gong2023multi, fu2022blindfl}, secret sharing (SS)\cite{li2023fedvs, huang2023efmvfl}, and hashing\cite{qiu2022all}. Employed to protect the privacy of individual embeddings in VFL, these methods, while effective, often lead to significant computation and communication overheads. Perturbation methods involve adding noise to embeddings to disturb potential reconstruction attacks, as detailed in~\cite{he2020attacking,wang2020hybrid, chen2020vafl}. Another variation treats noise as trainable parameters, aiming at minimizing privacy leakage without significant degradation of model performance, as explored in~\cite{mireshghallah2021not}. While these methods can mitigate data reconstruction attacks, they might reduce model accuracy significantly. Adversarial training techniques, as discussed in~\cite{sun2021defending, vepakomma2020nopeek}, train models in such a way that embeddings are less likely to leak raw features information. Although these methods are effective at reducing features leakage, they increase the computational burden and may affect the pefromance of the intended learning tasks.

\noindent\textbf{Detections against malicious attacks:} As malicious attacks such as FSHA and AGN substitute the intended training task, causing significant private features leakage, recent advancements in detection methods, notably SplitGuard~\cite{erdogan2022splitguard} and Gradient Scrutinizer~\cite{fufocusing}, have been developed to specifically address these types of attacks. SplitGuard operates by requiring passive clients to intermittently replace labels in a batch with randomized labels, referred to as fake batches, during training. The idea behind SplitGuard is that if the top model is genuinely engaged in learning the intended task, the angles and distances between the gradients from fake and regular batches will differ significantly from those between two regular batches.
Consequently, passive clients can detect anomalies by analyzing these differences in gradients. Unlike SplitGuard, Gradient Scrutinizer does not necessitate changes to labels. It operates by collecting and analyzing received batch gradients with identical and differing labels during each training round. The key finding is that during honest training, the gradient distances for embeddings with the different label should be greater than those for embeddings with same labels. By comparing these gradient distances, Gradient Scrutinizer can effectively detect deviations from honest training patterns, indicating potential attacks.

\noindent\textbf{Differences between defenses and detections:} While defenses are designed to significantly mitigate the privacy leakage by manipulating the embeddings, i.e., the embeddings are modified to preserve its representation ability as well as the privacy, the detection methods will not change the learning process of embeddings and only observe the variations of exchanged information to detect an attack. Specifically, current detections for data reconstruction attack analyze the difference of gradients between the honest and malicious training to detect the attack. Once there is an attack, the victims can stop the training process to prevent further privacy leakage.

%% file: Preliminary.tex
\section{Problem Description}
\subsection{VFL protocol}
In a VFL system, $N$ passive clients collaborate with an active client to train and utilize a VFL model. Each passive client, denoted as $n \in [N]$, operates a bottom model $f_n(\cdot)$. The active client manages a top model $f_0(\cdot)$ and a bottom model $f_a(\cdot)$. The training data $D_{train}:= (\boldsymbol{x}_j)_{j = 1}^M$ is vertically partitioned among $N + 1$ clients. For each sample $j\in [M]$, the active client holds both partial features $\boldsymbol{x}_{j,a}$ and the label $y_j$; while each passive client $n$, $n\in [N]$, possesses a disjoint portion of the features $\boldsymbol{x}_{j,n}$.

In the training phase, for a selected batch $\mathcal{B}$, each passive client $n$ computes the corresponding embeddings $\boldsymbol{h}_{j,n} = f_n(\boldsymbol{x}_{j,n}), j\in\mathcal{B}$, and sends them to the active client. The active client processes its partial features to produce embeddings $\boldsymbol{h}_{j,a} = f_a(\boldsymbol{x}_{j,a}),j\in\mathcal{B}$. 
Upon concatenating all batch embeddings, $\boldsymbol{h}_j = [\boldsymbol{h}_{j,a},\boldsymbol{h}_{j,1}, \ldots,\boldsymbol{h}_{j,N}]$, the active client completes forward propagation through the top model.  Loss is computed as $L = \frac{1}{|\mathcal{B}|}\sum_{j\in \mathcal{B}}l(f_0(\boldsymbol{h}_j),y_j)$, for some loss function (e.g., cross entropy) $l(\boldsymbol{a},b)$ between prediction $\boldsymbol{a}$ and label $b$. Gradients on the embeddings are sent back to each client for updating their respective bottom models. This process iterates until all models converge.

In the inference phase, when the prediction on a test sample is requested, each client computes an embedding using its bottom model and sends it to the active client. The active client aggregates these embeddings and uses the top model to predict the label. This approach ensures that data privacy is maintained by sharing only the embeddings rather than the raw features, and also enhances computational efficiency by leveraging the split learning mechanism.

\subsection{Threat model}
\textbf{Adversary's capability and knowledge.} In our threat model, the active client is treated as a malicious adversary capable of deviating from the standard VFL training protocol. Specifically, this adversary can alter the training task and transmits malicious gradients to other clients, e.g., alter the classification task to a reconstruction task. 
  As the active client, the adversary controls a top model $f_0(\cdot)$, and accesses partial data features $\boldsymbol{x}_{j,a}, j\in [M]$, along with a bottom model $f_a(\cdot)$ (see Figure~\ref{fig:vfl}).

  Additionally, the adversary possesses 
  an auxiliary dataset, $D_{aux}:= (\boldsymbol{x}_{i}, y_{i})_{i = 1}^ {M_{aux}}$, which shares a similar distribution with the training data but remains distinct from it, i.e., $D_{aux}\cap D_{train} = \emptyset$. Each sample in the auxiliary dataset contains the complete set of features and the corresponding label. 
 This assumption is congruent with previous attacks against SL~\cite{gao2023pcat, zhu2023passive, pasquini2021unleashing} and presents a more realistic capability compared to those models that assume direct querying of passive clients’ bottom models. 
In practice, adversaries often have access to subsets or samples of target data, which are used for validation or other purposes~\cite{sheller2020federated}. In some cases, training data may also include a mix of public and private datasets~\cite{tan2022towards, zhao2024medical}, which makes this assumption more practical.

 
\noindent \textbf{Goal and metric.} The primary objective of the adversary is to reconstruct target clients' data features, denoted as $\boldsymbol{X}_t = (\boldsymbol{x}_{j,n})_{n\in \mathcal{T}_N, j\in \mathcal{T}_X}$, where $\mathcal{T}_N$ is the set of target clients with size $|\mathcal{T}_N|\leq N$, and $\mathcal{T}_X$ is the set of target samples. We denote the reconstructed features as $\Tilde{\boldsymbol{X}}_t$, and use the mean squared error (MSE) between $\boldsymbol{X}_t$ and $\Tilde{\boldsymbol{X}}_t$ over target samples as the metric to measure the effectiveness of the attack. 


%% file: challenge.tex
\section{Challenges} 
In this section, we point out the main challenges to achieve effective and stealthy data reconstruction attack on VFL. We also review the current attacks' limitations.

\noindent\textbf{Challenge 1: distributed features and limited model access.} VFL involves multiple clients each holding a unique subset of data features and a bottom model. 
This dispersed and heterogeneous 
environment raises challenges for the adversary, to use its own partial features to effectively reconstruct target features from various sources. Additionally, given the distributed nature of bottom models across multiple clients, each guarding its data privacy, any attempt at querying or data injection would trigger an alarm for potential privacy breach and thus, is typically prohibited. This inaccessibility of passive clients' models complicates the adversary's strategy to manipulate them for data leakage.

\begin{figure}[htbp]
    \centering
    \includegraphics[width=0.8\linewidth]{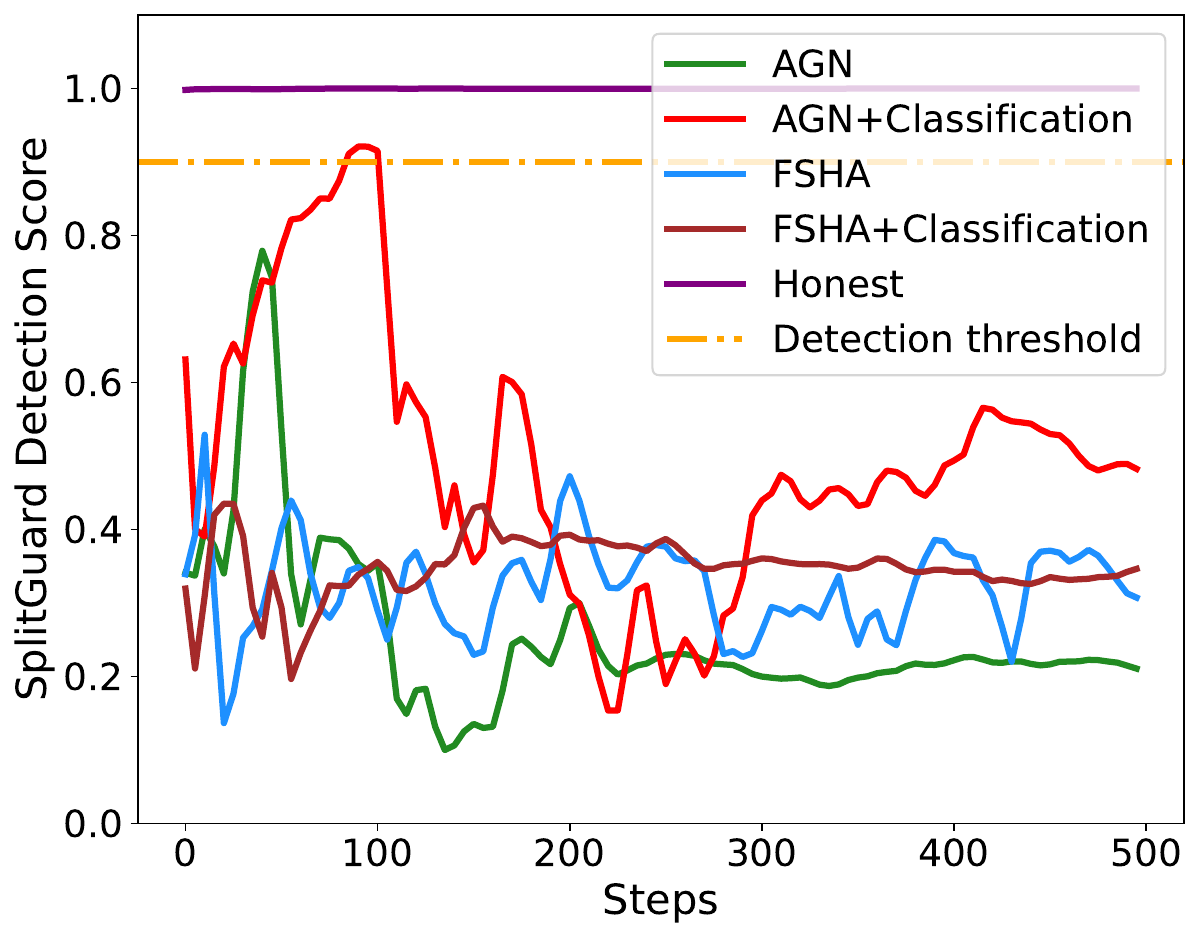}
    \caption{Comparison of Splitguard scores for two malicious attacks, and their variants with modified loss functions.}
    \label{fig:def_back}
\end{figure}

\noindent\textbf{Challenge 2: powerful detection strategies.} 
Malicious data reconstruction attacks are prone to detection by sophisticated mechanisms such as SplitGuard and Gradient Scrutinizer. One major reason is that current malicious attacks, e.g. AGN and FSHA, alter the honest training task, i.e., replace the honest training loss function with malicious loss function, and thus affecting the transmitted gradients. While such detections are highly sensitive to deviations in training behavior, as they typically monitor gradient flows for anomalies that could indicate malicious activities. 
To make the attack more stealthy, a natural solution is to integrate the honest training task (e.g., classification) into the malicious task. To evaluate the efficacy of this solution, we employed SplitGuard as a detection mechanism to test against modified malicious attack strategies AGN and FSHA on MNIST dataset, where the training loss is the summation of classifcation loss and the malicious loss. 
As depicted in Figure~\ref{fig:def_back}, the detection scores for both attacks, even with the modified loss, remained below the threshold of 0.9, yielding successful detection. This outcome suggests that simply combining loss functions is insufficient to circumvent current detections.

Existing HBC data reconstruction attacks on VFL operates under strong, often unrealistic assumptions about adversary's capabilities, such as the ability to query the target model or inject data directly into target clients, which are impractical in real-world scenarios. 
 Malicious attacks like FSHA and AGN completely ignore the label information in their designs, which is the root cause of their detectability, and incorporating the labels into attack designs can potentially further improve the reconstruction performance.



The above challenges steer us toward a compelling research question: \textit{"Under a practical adversary model, can we develop a malicious data reconstruction attack in VFL that successfully circumvents current detection mechanisms while achieving superior reconstruction performance?"} We affirmatively answer this question by developing a novel attack strategy \sys, which addresses all highlighted challenges. 



%% file: methodology.tex
\section{Methodology of URVFL}
\begin{figure*}[ht!] 
  \centering
  \includegraphics[width=\textwidth]{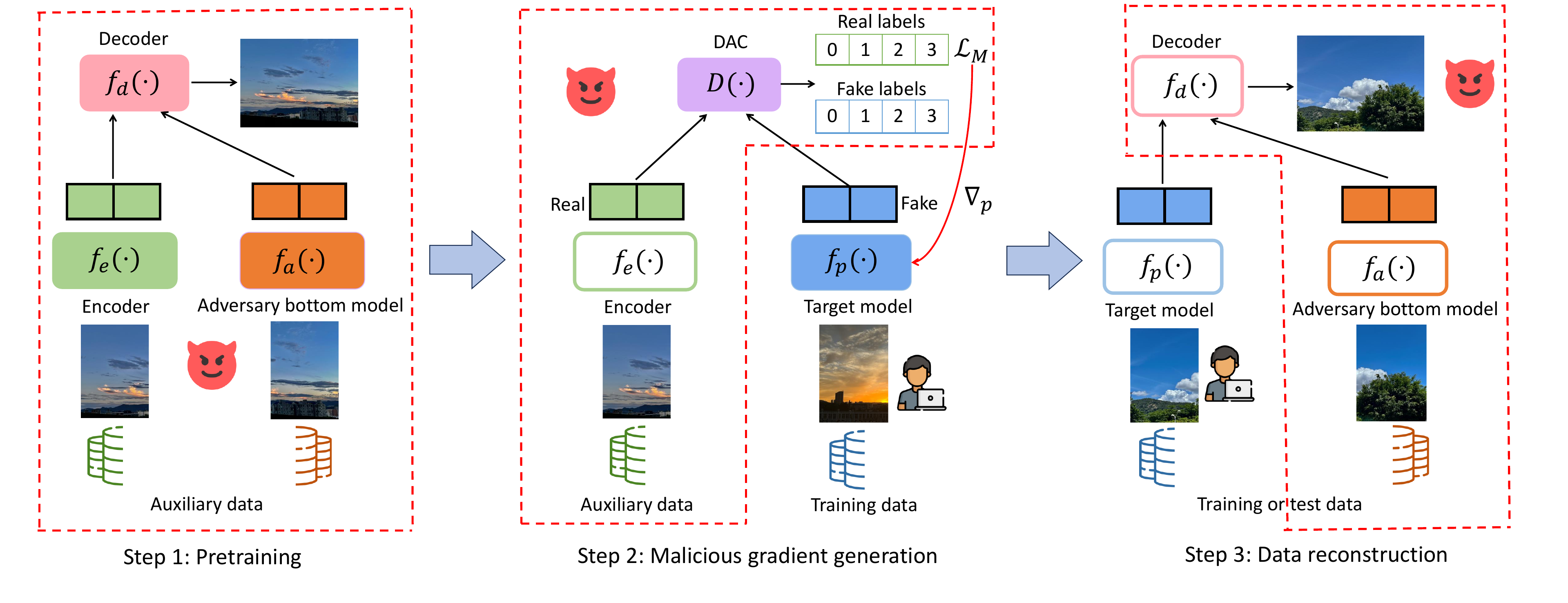} 
  \vspace{-8mm}
  \caption{Workflow of URVFL. The opaque rectangle indicates that the model is being trained, while the transparent rectangle represents that the model is frozen and only utilized in forward propagation. }
  \label{fig:dmavfl_flow}
\end{figure*}

In this section, we introduce our \sys strategy. As illustrated in Figure~\ref{fig:dmavfl_flow}, \sys consists of the following three steps. 
\begin{enumerate}[leftmargin=*]
    \item \textbf{Pretraining:} Before initiating VFL training, the adversary sets up three local models consisting of an encoder, a decoder, and a Discriminator with Auxiliary Classifier (DAC). The pretraining step involves the concurrent training of these models. This step is critical for minimizing reconstruction loss and effectively preparing the models for the subsequent steps of the attack.

\item \textbf{Malicious gradient generation:} With the onset of VFL training, the adversary freezes the encoder to preserve its learned embeddings and replaces the conventional top model with the DAC. The DAC is instrumental in transferring the embedding distribution from the encoder into the target model and integrating label information through classification processes. This integration ensures that the malicious training is indistinguishable from honest training, enhancing the stealthiness of the attack. Utilizing the malicious loss computed through DAC, the adversary meticulously crafts malicious gradients. These gradients are then dispatched to the target clients to manipulate their models, mimicking the embedding distribution of the encoder.

\item \textbf{Data reconstruction:} In the data reconstruction phase, the adversary leverages the decoder to reconstruct private features of target clients, from embeddings uploaded by passive clients and adversary's local embedding. 
\end{enumerate}

\subsection{Setup}
We first denote a target model $f_{p}(\cdot) := (f_n(\cdot))_{n \in [N]}$ as the collection of the bottom models of all passive clients. In the setup phase, the adversary prepares the following three networks in replacement of the original top model: 
\begin{enumerate}[leftmargin=*]
    \item Encoder ($f_e(\cdot)$): The encoder $f_e(\cdot)$ is designed to mimic the functionality of the target model $f_{p}(\cdot)$. 
    The dimensions of the encoder are carefully aligned with those of $f_{p}(\cdot)$, such that its input dimension equals to the total number of features on the passive clients, and its output dimension equals to the sum of output dimensions of all passive clients' the bottom models.  
    
    \item Decoder ($f_d(\cdot)$): The decoder is tailored to reconstruct the target clients' data features. It accepts concatenated embeddings as input and outputs data features of the target clients.
    \item Discriminator with Auxiliary Classifier (DAC) ($D(\cdot)$) : The DAC is engineered to classify and discriminate between real labels $y^+$ and fake labels $y^-$, aiding in the validation of the embeddings' authenticity and malicious gradient generation.
\end{enumerate}

Additionally, the adversary maintains a bottom model $f_a(\cdot)$. This model processes the partial data features available to the adversary, capturing their correlations with the target features.

\subsection{Pretraining}

Prior to commencing VFL training, the adversary constructs an encoder-decoder structure consisting of the adversary's bottom model $f_a(\cdot)$, the encoder $f_e(\cdot)$, and the decoder $f_d(\cdot)$. This setup is pretrained on the auxiliary dataset $D_{aux}$. The objective of this pretraining phase is threefold: to train the decoder to reconstruct target features, to enable the adversary's bottom model to learn the relationship between target and adversary's partial features, and to allow the encoder to capture the representation of passive clients' features for effective reconstruction.

 In each iteration of training, the adversary selects a batch of auxiliary data, denoted as $\mathcal{B}_{aux}$. This batch is partitioned into two subsets: the adversary's features  $\boldsymbol{x}_{i,a}$ and the passive features $\boldsymbol{x}_{i,p}, i\in \mathcal{B}_{aux}$, which are fed into $f_a(\cdot)$ and $f_e(\cdot)$, respectively. The resulting embeddings, $\boldsymbol{h}_{i,a} = f_a(\boldsymbol{x}_{i,a})$ and $\boldsymbol{h}_{i,p} = f_e(\boldsymbol{x}_{i,p}), i\in \mathcal{B}_{aux}$ are concatenated to form $\boldsymbol{h}_i = (\boldsymbol{h}_{i,a} \| \boldsymbol{h}_{i,p})$. These concatenated embeddings are then forwarded to the decoder, which produces synthetic target features $\Tilde{\boldsymbol{x}}_{i,t} = f_d(\boldsymbol{h}_i)$. The primary objective of the adversary is to minimize the reconstruction loss, which is computed as follows:
\begin{equation}
    \mathcal{L}_R = \frac{1}{|\mathcal{B}_{aux}|}\sum_{i\in \mathcal{B}_{aux} }MSE(\Tilde{\boldsymbol{x}}_{i,t}, \boldsymbol{x}_{i,t}).\label{LR}
\end{equation}
    
\begin{algorithm}[htb]

\KwInput{Auxiliary dataset $D_{aux}= \{\boldsymbol{x}_{i,a}, \boldsymbol{x}_{i,p}\}_{i=1}^{M_{aux}}$. }

\KwOutput{Encoder $f_e(\cdot)$, decoder $f_d(\cdot)$, adversary's bottom model $f_a(\cdot)$.}

\While{$f_e(\cdot)$, $f_d(\cdot)$, and $f_a(\cdot)$ not converge}
{

The adversary select a batch $\mathcal{B}_{aux}$ from $D_{aux}$;

Compute $\boldsymbol{h}_{i,a} = f_a(\boldsymbol{x}_{i,a}),i \in \mathcal{B}_{aux}$, and  $\boldsymbol{h}_{i,p} = f_e(\boldsymbol{x}_{i,p}), i \in \mathcal{B}_{aux}$;

Compute $\Tilde{\boldsymbol{x}}_{i, t} = f_d([\boldsymbol{h}_{i,a}\| \boldsymbol{h}_{i,p}]), i\in\mathcal{B}_{aux}$;

$\mathcal{L}_R \gets \frac{1}{|\mathcal{B}_{aux}|}\sum_{i\in \mathcal{B}_{aux} }MSE(\Tilde{\boldsymbol{x}}_{i, t}, \boldsymbol{x}_{i, t})$;

$\boldsymbol{\nabla}_e \gets$ Gradient($\mathcal{L}_R, f_e(\cdot)$);

$\boldsymbol{\nabla}_a \gets$ Gradient($\mathcal{L}_R, f_a(\cdot)$);

$\boldsymbol{\nabla}_d \gets$ Gradient($\mathcal{L}_R, f_d(\cdot)$);

$f_e(\cdot) \gets \text{Model\_update}(f_e(\cdot), \boldsymbol{\nabla}_e $);

$f_a(\cdot) \gets \text{Model\_update}(f_a(\cdot), \boldsymbol{\nabla}_a $);

$f_d(\cdot) \gets \text{Model\_update}(f_d(\cdot), \boldsymbol{\nabla}_d $);

}
\caption{Pretraining procedure.}\label{alg1}
\end{algorithm}
The pretraining process is detailed in Algorithm~\ref{alg1}, where the function Gradient($L, \boldsymbol{\theta}$) computes the gradients of the loss $L$ with respect to parameters $\boldsymbol{\theta}$; and the function Model\_update($f(\cdot), \nabla_{\boldsymbol{\theta}}$) represents the operation of updating the model $f(\cdot)$ using gradients $\nabla_{\boldsymbol{\theta}}$.

\subsection{Malicious gradient generation}

After the pretraining, the adversary intends to transfer the  embeddings distribution from the encoder $f_e(\cdot)$ to the target model $f_p(\cdot)$. Since the adversary lacks direct access to modify $f_p(\cdot)$, they resort to transmitting malicious gradients to the passive clients, which are designed to guide $f_p(\cdot)$ to mimic the behavior of $f_e(\cdot)$. In this sequel, for ease of exposition, we describe all operations with $f_p(\cdot)$, which means that the same operation is simultaneously performed on the bottom models of all passive clients.

An intuitive solution involves employing a discriminator to distinguish between embeddings from the encoder (labeled as real) and those produced by the target model (labeled as fake). The target model is then trained to deceive the discriminator into recognizing its embeddings as real, thus aligning its distribution with that of the encoder. This strategy focuses on minimizing the Jensen–Shannon divergence (JS)~\cite{fuglede2004jensen} between the distributions of the embeddings from the auxiliary and training data, i.e., $\textup{JS}(P_{\boldsymbol{H}_{aux}}\| P_{\boldsymbol{H}_{train}})$, which facilitates the transfer of embeddings distribution. 

However, a primary concern is that the discriminator approach completely ignores the label information available at the adversary, which may lead to 1) sub-optimal reconstruction performance; and 2) detection by methods that can tell the difference between gradients computed using and without using labels.
To address these issues, we employ DAC, $D(\cdot)$, defined in (\ref{DAC_def}), which not only differentiates between real and fake embeddings but also distinguishes the corresponding labels.
\begin{equation}
    \min_D \mathbb{E}_{\boldsymbol{h},y \sim P_{\boldsymbol{H}_{aux}, Y}} CE(y^{+}, D(\boldsymbol{h})) + \mathbb{E}_{\boldsymbol{h},y \sim P_{\boldsymbol{H}_{train}, Y}} CE(y^{-}, D(\boldsymbol{h})).
\label{DAC_def}
\end{equation}
Here, $P_{\boldsymbol{H}_{aux}, Y}$ and $P_{\boldsymbol{H}_{train}, Y}$ represent the joint distributions of embeddings and labels from the auxiliary and training datasets, respectively. The labels $y^+$ and $y^-$ denote real and fake label $y$, respectively, and $\text{CE}(\cdot)$ represents the cross-entropy loss.

To align the target model's feature embeddings more closely with the encoder's embedding, the adversary modifies the honest training task into an adversarial task: 
\begin{equation}
    \min_{\boldsymbol{h}} \mathbb{E}_{\boldsymbol{h},y \sim P_{\boldsymbol{H}_{train}, Y}} CE(y^{+}, D(\boldsymbol{h})).
\end{equation}
Incorporating DAC with this adversarial task minimizes the Kullback–Leibler (KL) divergence between $P_{\boldsymbol{H}_{aux}, Y}$ and $P_{\boldsymbol{H}_{train}, Y}$,
i.e., $\textup{KL}(P_{\boldsymbol{H}_{aux}, Y}\| P_{\boldsymbol{H}_{train}, Y})$. This integration ensures that the adversarial task remains indistinguishable from the honest training task. The correlation of generated gradients with labels not only reduces the detectability, but also enhances the effectiveness of the embedding distribution transfer and thus the reconstruction performance.

\noindent\textbf{Comparison between discriminator and DAC:} We carry out empirical studies on the distribution transfer capabilities of discriminator and the DAC, via experiments on the Credit dataset with 2 labels, and the MNIST dataset with 10 labels.  
We use t-distributed stochastic neighbor embedding (t-SNE) to visualize the embedding distributions,
and quantify the alignment of the target model’s embeddings with those of the encoder by measuring the average cosine distance between them.

\begin{figure*}[htbp]
    \centering
    \begin{subfigure}[t]{0.32\textwidth}
        \centering
        \includegraphics[width=\linewidth]{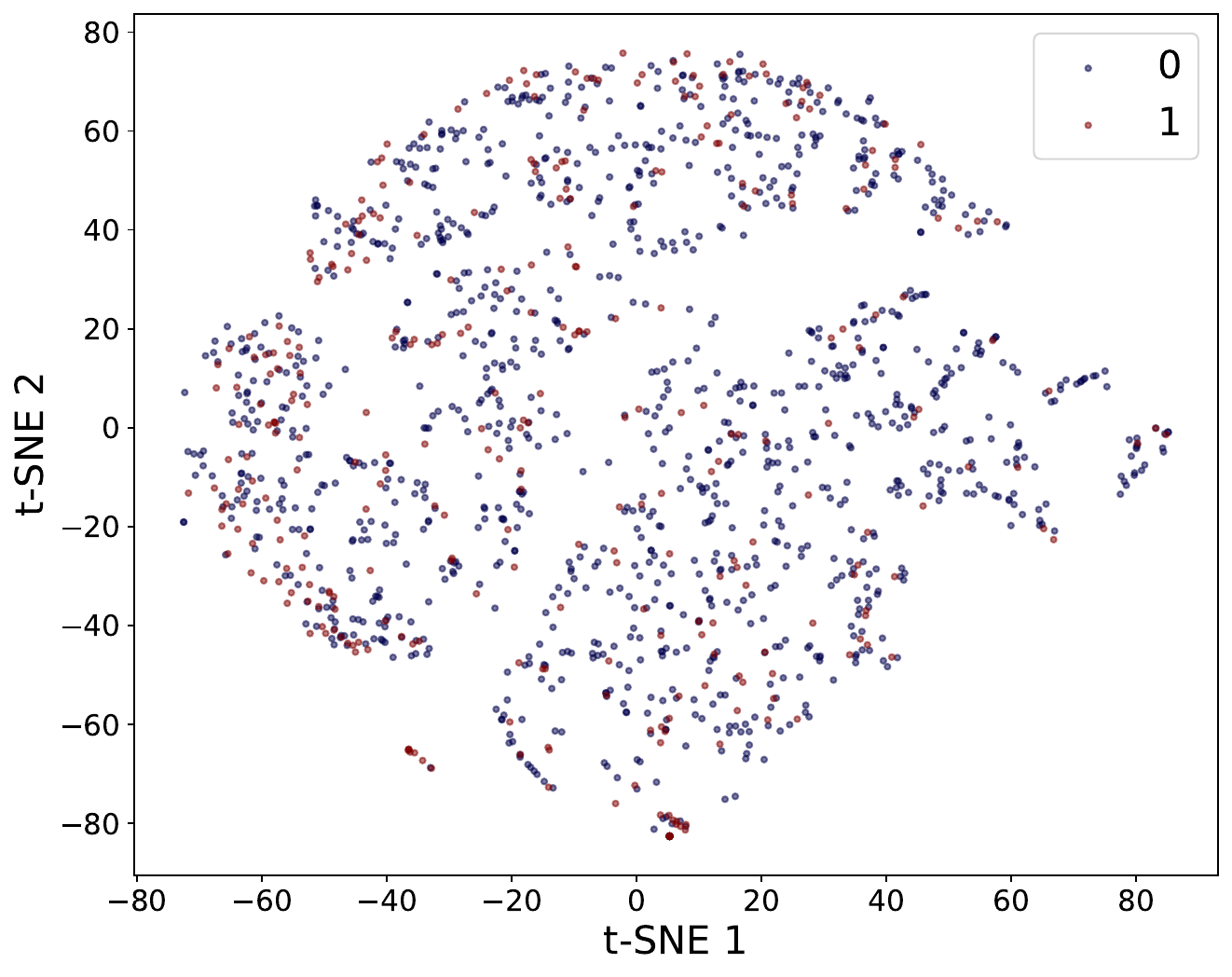}
        \captionsetup{justification=centering, width=.8\linewidth}
        \caption{Encoder's embedding distribution.}
        \label{fig:sub1}
    \end{subfigure}\hfill
    \begin{subfigure}[t]{0.32\textwidth}
        \centering
        \includegraphics[width=\linewidth]{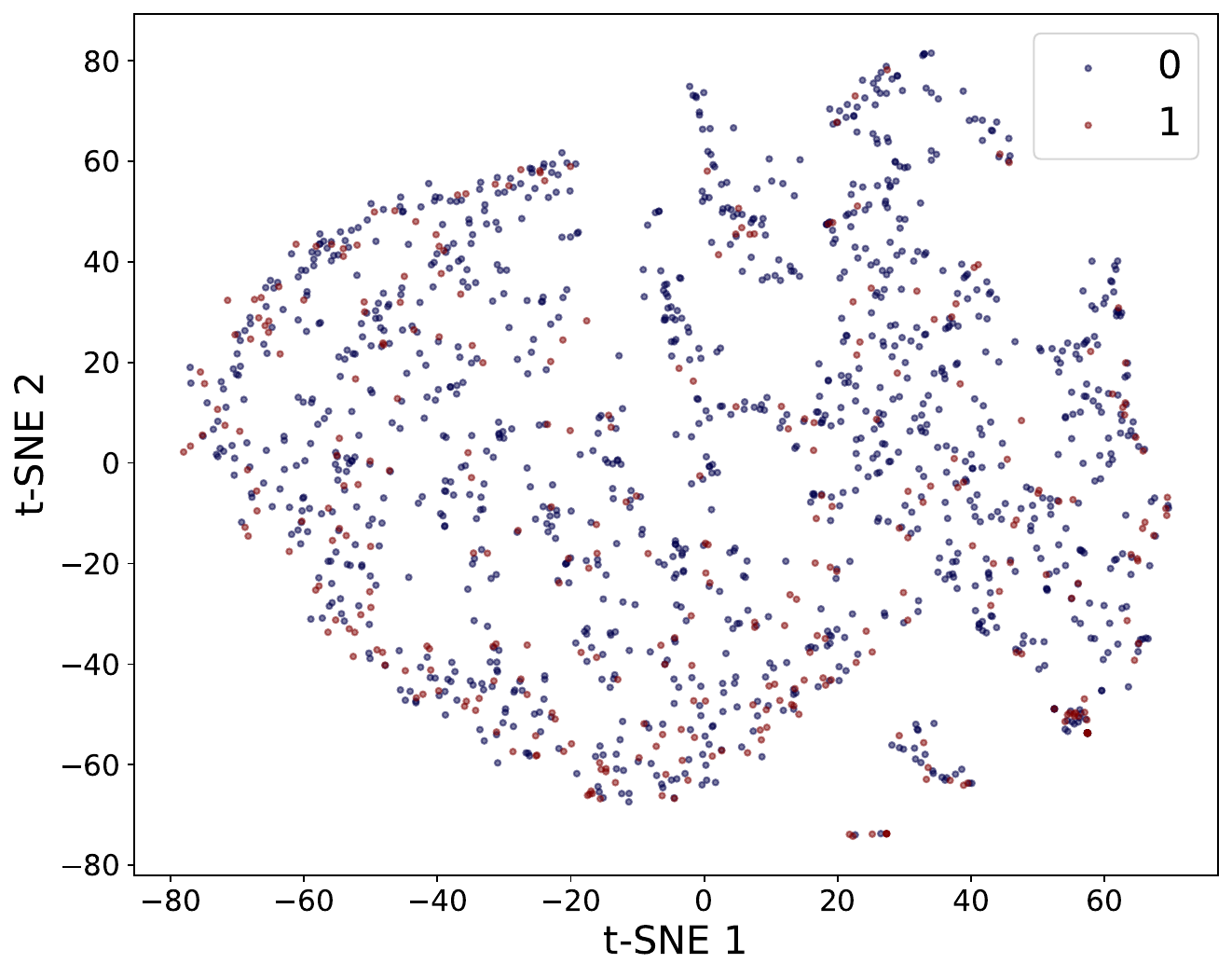}
        \captionsetup{justification=centering, width=.8\linewidth}
        \caption{Target model's embedding distribution using discriminator. Average cosine distance is 0.6121.}
        \label{fig:sub2}
    \end{subfigure}\hfill
    \begin{subfigure}[t]{0.32\textwidth}
        \centering
        \includegraphics[width=\linewidth]{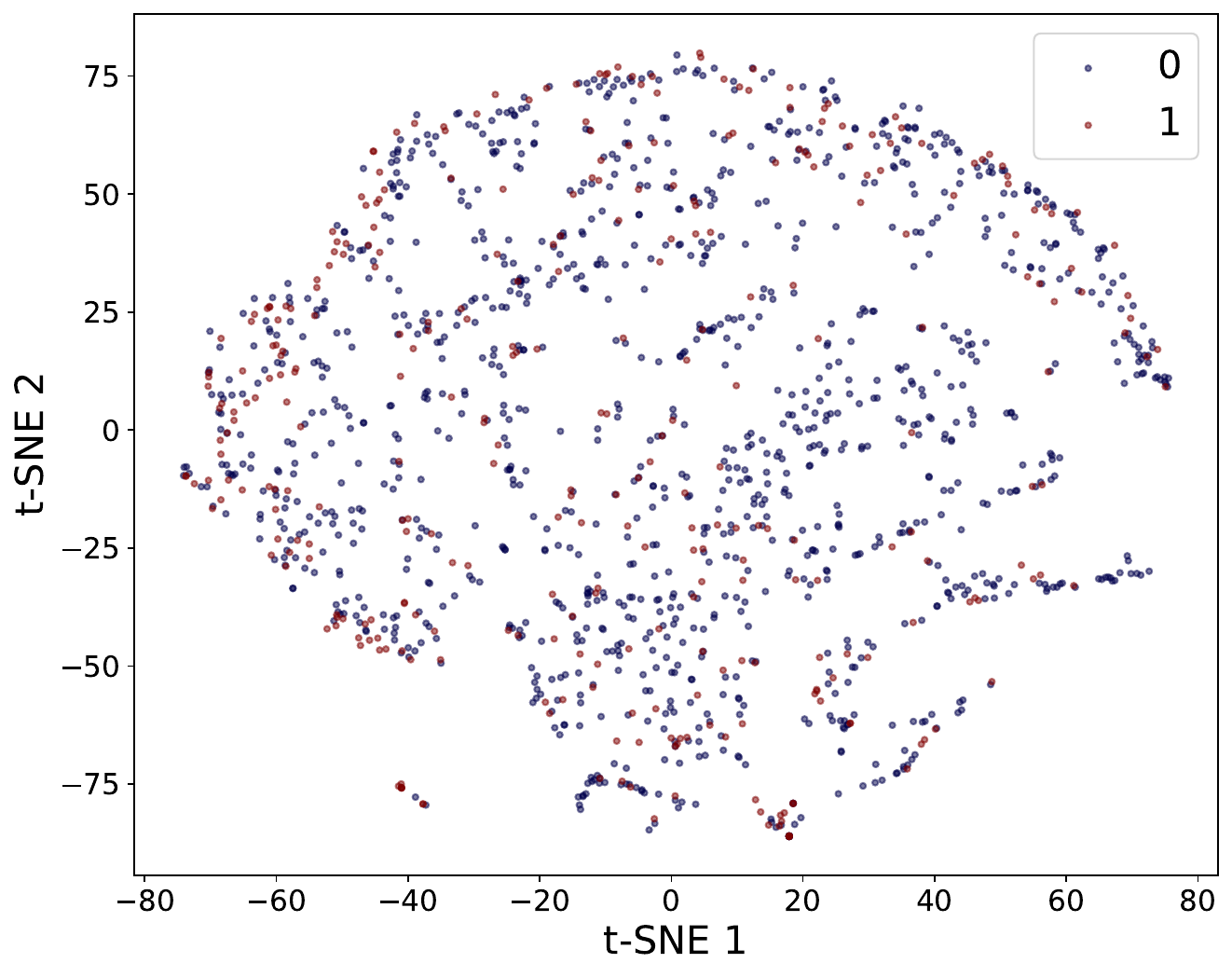}
        \captionsetup{justification=centering, width=.8\linewidth}
        \caption{Target model's embedding distribution using DAC. Average cosine distance is 0.1373.}
        \label{fig:sub3}
    \end{subfigure}
    \caption{t-SNE visualization on Credit dataset with 2 classes.}
    \label{fig:DAC_credit}
\end{figure*}

\begin{figure*}[htbp]
    \centering
    \begin{subfigure}[t]{0.32\textwidth}
        \centering
        \includegraphics[width=\linewidth]{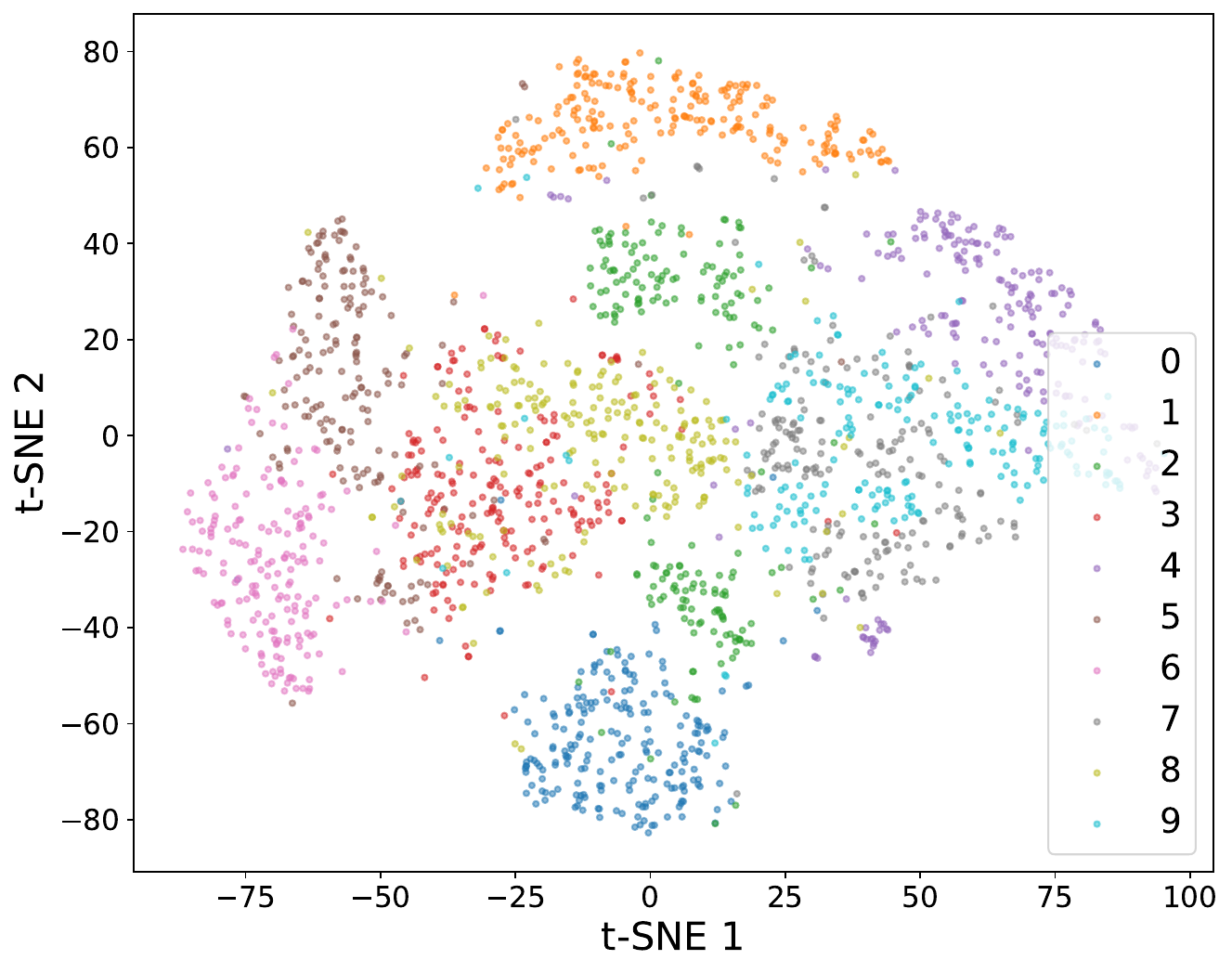}
        \captionsetup{justification=centering, width=.8\linewidth}
        \caption{Encoder's embedding distribution.}
        \label{fig:sub4}
    \end{subfigure}\hfill
    \begin{subfigure}[t]{0.32\textwidth}
        \centering
        \includegraphics[width=\linewidth]{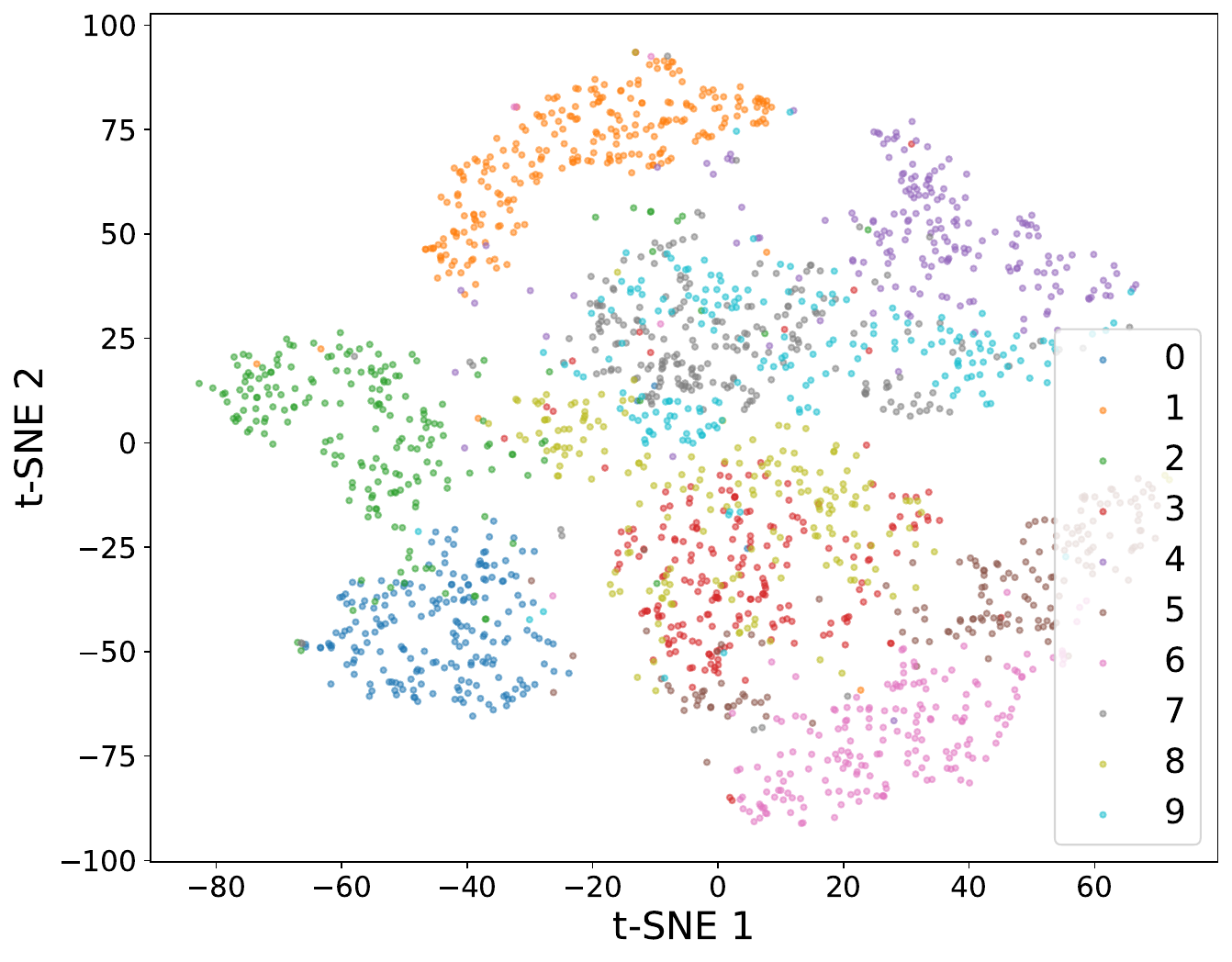}
\captionsetup{justification=centering, width=.8\linewidth}
        \caption{Target model's embedding distribution using discriminator. Average cosine distance is 0.0412.}
        \label{fig:sub5}
    \end{subfigure}\hfill
    \begin{subfigure}[t]{0.32\textwidth}
        \centering
        \includegraphics[width=\linewidth]{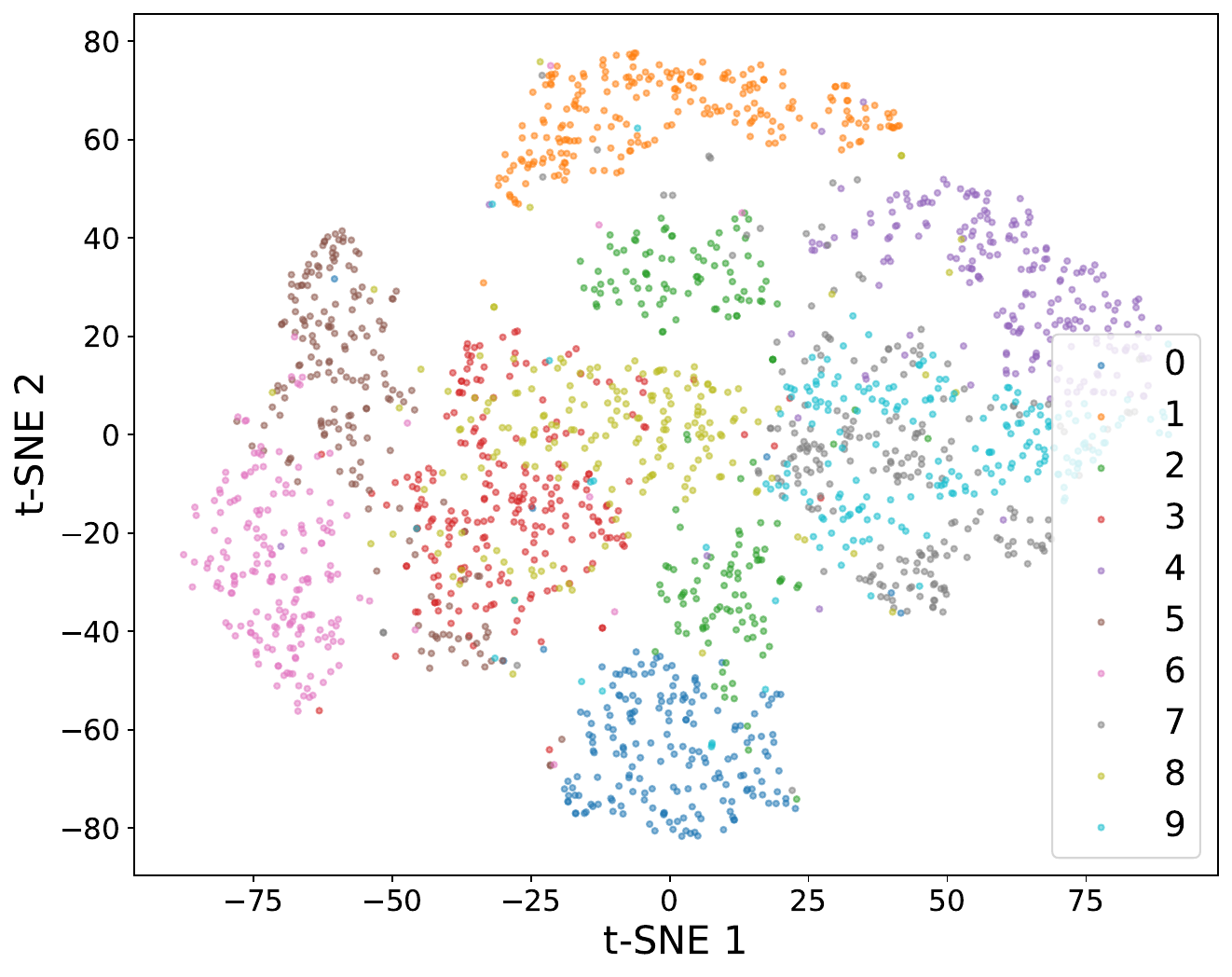}
        \captionsetup{justification=centering, width=.8\linewidth}
        \caption{Target model's embedding distribution using DAC. Average cosine distance is 0.0283.}
        \label{fig:sub6}
    \end{subfigure}
    \caption{t-SNE visualization on MNIST dataset with 10 classes.}
    \label{fig:DAC_MNIST}
\end{figure*}

The results, illustrated in Figure~\ref{fig:DAC_credit} and \ref{fig:DAC_MNIST}, indicate a superior performance of the DAC in distribution transfer over discriminator.
The t-SNE visualizations reveal that the embeddings' distribution of the target model more closely resembles that of the encoder when the DAC is employed. Furthermore, a smaller average cosine distance between the embeddings of the target model and the encoder is observed when using the DAC. 


\noindent\textbf{Malicious gradient generation using DAC:} During each training round, both the adversary and other clients select a batch of training data, denoted as $\mathcal{B}_{train}$. 
Upon receiving the embeddings $\boldsymbol{h}_{j,p}, j\in \mathcal{B}_{train}$ from passive clients, the adversary computes the malicious loss defined by: 
\begin{equation}
    \mathcal{L}_{M} = \frac{1}{|\mathcal{B}_{train}| } \sum_{j\in \mathcal{B}_{train}}CE\left(y_j^+, D(\boldsymbol{h}_{j,p})\right).
\label{LM}\end{equation}

The adversary then calculates the gradient $\pdv{L_M}{\boldsymbol{h}_{j,p}}$ and transmits it back to the passive clients to guide their bottom (target) models' learning processes. On the other hand, the adversary selects a batch from the auxiliary data $\mathcal{B}_{aux}$ and computes the embeddings $\boldsymbol{h}_{i,p} = f_e(\boldsymbol{x}_{i,p}), i\in \mathcal{B}_{aux}$. These embeddings are used to compute the DAC loss:
\begin{equation}
\begin{aligned}
    \mathcal{L}_D &= \frac{1}{|\mathcal{B}_{train}|} \sum_{j\in \mathcal{B}_{train}}CE(y_j^{-}, D(\boldsymbol{h}_{j,p}))\\ &+ \frac{1}{|\mathcal{B}_{aux}|} \sum_{i\in \mathcal{B}_{aux}}CE(y_i^{+}, D(\boldsymbol{h}_{i,p})).\label{LD}
    \end{aligned}
\end{equation}
Minimizing $\mathcal{L}_D$ helps the DAC to better differentiate between real and fake embeddings associated with their labels, thereby refining its ability to guide the learning process of the target model while staying undetected. The detailed steps of malicious gradient generation are given in Algorithm~\ref{alg2}.

\begin{algorithm}[htb]

\KwInput{Auxiliary dataset $D_{aux}= \{\boldsymbol{x}_{i,a}, \boldsymbol{x}_{i,p}\}_{i=1}^{M_{aux}}$ and training dataset $D_{train} = \{\boldsymbol{x}_{j,a}, \boldsymbol{x}_{j,p}\}_{j=1}^{M}$.}

\KwOutput{Encoder $f_e(\cdot)$, DAC $D(\cdot)$, and target  model $f_p(\cdot)$.}

\textit{Adversary Procedure:}

\While{VFL training}
{

All clients agree a batch data $\mathcal{B}_{train}$ from $D_{train}$;

Receive and record passive clients' embeddings $\boldsymbol{h}_{j,p},j \in \mathcal{B}_{train}$;

Compute the loss $\mathcal{L}_M = \frac{1}{|\mathcal{B}_{train}|}\sum_{j\in\mathcal{B}_{train}}CE(y_j^{+}, D(\boldsymbol{h}_{j,p}))$;

$\boldsymbol{\nabla}_{p}  \gets $Gradient($\mathcal{L}_M, \boldsymbol{h}_{j,p}, j \in \mathcal{B}_{train})$;

Send $\boldsymbol{\nabla}_{p}$ to the passive clients;

The adversary select a batch of data $\mathcal{B}_{aux}$ from $D_{aux}$;

Compute $\boldsymbol{h}_{i,p} = f_e(x_{i,p}), i \in \mathcal{B}_{aux}$;

Compute the DAC loss $\mathcal{L}_D$ in (\ref{LD});

$\boldsymbol{\nabla}_{D}  \gets$  Gradient($\mathcal{L}_D, D(\cdot))$;

$D(\cdot) \gets \text{Model\_update}(D(\cdot), \boldsymbol{\nabla}_D $);
}

\textit{Passive Clients Procedure:}

\While{VFL training }{

All clients agree a batch data $\mathcal{B}_{train}$ from $D_{train}$;

Compute embeddings $\boldsymbol{h}_{j,p} = f_p(\boldsymbol{x}_{j,p}), j\in \mathcal{B}_{train}$ and send embeddings to the active client;

Receive gradient $\boldsymbol{\nabla}_p$;

$f_p(\cdot) \gets \text{Model\_update}(f_p(\cdot), \boldsymbol{\nabla}_p)$;
}

\caption{Malicious gradient generation.}\label{alg2}
\end{algorithm}
\noindent \textbf{An alternative synchronous training strategy: URVFL\_sync.} We also consider a variant of \sys, named URVFL\_sync, that replaces the pretraining with synchronous training of all models when generating malicious gradients. Specifically, during each training round, the adversary selects a batch of auxiliary data, $\mathcal{B}_{aux}$. The models $f_a(\cdot), f_d(\cdot)$ and $ f_e(\cdot)$ are trained on $\mathcal{B}_{aux}$ to minimize the reconstruction loss $\mathcal{L}_R$ in~(\ref{LR}). Upon receiving the embeddings $\boldsymbol{h}_{j,p}, j\in \mathcal{B}_{train}$ from the passive clients, the adversary calculates the gradients $\pdv{L_M}{\boldsymbol{h}_{j,p}}$ by minimizing the malicious loss $\mathcal{L}_M$ in~(\ref{LM}), and transmits these gradients back to the passive clients. Following this, the adversary computes the embedding $\boldsymbol{h}_{i,p}, i\in \mathcal{B}_{aux}$ using the updated encoder to calculate the DAC loss $\mathcal{L}_D$ in~(\ref{LD}), and finally updates $D(\cdot)$.
 This process is repeated until all models reach convergence. The detailed steps of URVFL\_sync, as outlined in Algorithm~\ref{alg3}, are shown in Appendix~\ref{appB}. Subsequent sections will demonstrate that URVFL\_sync exhibits better reconstruction performance on certain datasets. 

\subsection{Data reconstruction}
Having maliciously guided the training of the target model $f_p(\cdot)$ to mimic the encoder $f_e(\cdot)$, the adversary leverages the trained decoder $f_d(\cdot)$, alongside its bottom model $f_a(\cdot)$ to perform data reconstruction. For a target sample $\boldsymbol{x}$ (from training or test set) with adversary's partial features $\boldsymbol{x}_a$ and passive clients' features $\boldsymbol{x}_p$, the adversary can reconstruct the private target features $\Tilde{\boldsymbol{x}}_t = f_d(f_a(\boldsymbol{x}_a)\| f_p(\boldsymbol{x}_p))$. 

%% file: experiment.tex
\section{Experiments}


\subsection{Experiment setup}

\textbf{Datasets and data processing:} We conduct experiments across five representative datasets for VFL tasks, including two tabular datasets (i.e., Credit~\cite{misc_default_of_credit_card_clients_350} and RT\_IOT2022~\cite{misc_rt-iot2022__942}) and three image datasets (i.e., MNIST~\cite{deng2012mnist}, CIFAR-10~\cite{krizhevsky2009learning}, and Tiny imagenet~\cite{le2015tiny}). Each dataset is partitioned into auxiliary, training, and test sets. For MNIST and CIFAR-10, we use the standard test sets provided. For the Credit dataset, we randomly allocate 30\% of the samples to the test set, and for RT\_IOT2022 and Tiny imagenet, 20\% of the samples are selected as the test set. The auxiliary set, chosen randomly from the original training set, comprises 10\% of the training set size in all main experiments, ensuring $|D_{aux}|: |D_{train}| = 1 : 10$ and $D_{aux}\cap D_{train} = \emptyset$. For image dataset, we normalize the data features to the range [-1,1]. For tabular dataset, most features are continuous or integer. We encode the categorical feature in Credit dataset using sklearn~\cite{pedregosa2011scikit} and normalize all features with StandardScaler. Data features are evenly split between passive (target) and active clients, with each holding 50\% of the features. 

\noindent\textbf{Models:} 
The adversary's encoder $f_e(\cdot)$ is configured identically to the passive clients' bottom model $f_p(\cdot)$ across all datasets. The structure of the adversary's bottom model $f_a(\cdot)$ mirrors that of $f_p(\cdot)$. For tabular data, we employ multi-layer perceptron (MLP) models for $f_e(\cdot), f_a(\cdot)$, $f_p(\cdot)$, the decoder $f_d(\cdot)$, and the DAC $D(\cdot)$. For image data, convolutional neural networks (CNNs) serve as the foundational architecture, with the CIFAR-10 and Tiny imagenet models featuring a residual block composed of three CNN layers and a skip connection. Detailed dataset information, model structures, and training parameters are provided in the Appendix~\ref{appendix_a}.

\noindent\textbf{Metrics:} We measure the reconstruction performance using average MSE between the reconstructed features and the original features, denoted as \textit{Recon MSE}. Besides, to comprehensively measure the reconstruction performance on image datasets, we also leverage \textit{PSNR} and \textit{SSIM} as the metrics. To analyze the effectiveness of DAC compared to other embedding distribution transfer techniques (e.g., using discriminator or the top model), we measure both the average MSE and the cosine distance between the encoder's (or shadow model's) embeddings and the target model's embeddings, denoted as \textit{Emb MSE} and \textit{Emb Cos}, respectively. Before computing the cosine distance, all embeddings are normalized. 

\noindent\textbf{Environment:} All experiments are conducted on a single machine equipped with four NVIDIA RTX 4090 GPUs. Each experiment is repeated five times, with the average results reported.

\begin{table*}[tb]
\centering
\caption{Results of data reconstruction attacks on tabular datasets. Attack methods followed by SG or GS represent implementations under SG or GS detections. \sys and URVFL\_sync achieve the same performance with and without detections. Some entries are missing due to 1) some attacks do not use encoders for data reconstruction, and hence do not have embedding MSE and cosine distance; and 2) the GS defense does not work with Credit dataset.
}
\scriptsize
\begin{tabular}{lccccccc
}
\toprule
Method & \multicolumn{3}{c}{Credit} & \multicolumn{3}{c}{RT\_IOT2022} \\ 
\cmidrule(lr){2-4} \cmidrule(lr){5-7} 
 & Recon MSE & Emb MSE & Emb Cos & \begin{tabular}[c]{@{}c@{}}Recon\\ MSE\end{tabular} & \begin{tabular}[c]{@{}c@{}}Emb \\ MSE \end{tabular} & \begin{tabular}[c]{@{}c@{}} Emb \\ Cos\end{tabular} \\
\midrule
GRNA & $1.1822\pm0.0541$ & {-} & {-} & $2.2542\pm0.0325$& {-} & {-}  \\
GIA & $0.9056\pm0.0002$ & $\textbf{0.2738}\pm0.0008$ & $0.4681\pm0.0215$ & $1.8535\pm0.0012$ & $1.7276\pm0.2356$ & $0.4491\pm 0.0371$ \\
AGN & $0.9656\pm0.1196$ & {-} & {-} & $2.2311\pm0.1256$ & {-} & {-}\\
AGN-SG & $1.4155\pm 0.1734$ & {-} & {-} & $ 2.3967\pm 0.2628$ & {-} & {-}  \\
AGN-GS& {-} & {-} & {-} & $ 2.5670\pm 0.2909$ & {-} & {-}  \\
PCAT & $0.8612 \pm 0.0984$ & $0.5282 \pm 0.0467$ & $0.5486 \pm 0.0197$ & $2.2963\pm0.4388$ & $\textbf{1.5646}\pm0.2488$ & $0.3742\pm0.0195$  \\
SDAR & $0.5327\pm 0.0374$ & $0.5451\pm 0.1645$ & $0.3103\pm 0.0931$ & $1.6084\pm0.1604 $& $2.9362 \pm 1.3356$ & $0.2281\pm 0.1689$ \\
FSHA & $0.5032 \pm 0.0382$ & $0.3906 \pm 0.0622$ & $0.2234 \pm0.0564$ & $1.5773\pm0.0507$ & $2.2990 \pm 0.6506$ & $0.4859 \pm  0.0984$  \\
FSHA-SG & $0.7884\pm0.1139$ & $0.6056\pm0.0704$ & $0.5640\pm0.0369$ & $ 1.6895\pm0.0419$ & $ 2.7922\pm1.0491$ & $0.5909\pm0.0745$  \\
FSHA-GS & {-} & {-} & {-} & $1.7442\pm0.0784$ & $ 2.5768\pm0.2169$ & $0.5534\pm0.1453$  \\
\textbf{URVFL (SG/GS)} & \textbf{$\textbf{0.4191} \pm 0.0541$} & $0.3078 \pm 0.0778$ & \textbf{$\textbf{0.1658}\pm0.0551$} & $1.3821\pm0.0244$ & $1.7894\pm 0.7435$ & $\textbf{0.0549}\pm 0.0113$  \\
\textbf{URVFL\_sync (SG/GS) }& $0.4722 \pm0.0228$ & $0.3720 \pm0.0462$ &$0.2277\pm0.0441$ & $\textbf{1.3277}\pm0.0665$ & $2.3372\pm0.5489$ & $0.0938\pm0.0340$  \\
\bottomrule
\end{tabular}\label{tab:main1}
\vspace{-2mm}
\end{table*}

\begin{table*}[tb]
\centering
\caption{Results of data reconstruction on image datasets. Attack methods followed by SG or GS represent implementations under SG or GS detections. \sys and URVFL\_sync achieve the same performance with and without defenses. Some entries are missing since these attacks do not use encoders for data reconstruction, and hence, there is no embedding distance.
}
\begin{adjustbox}{width=\textwidth,center}
\begin{tabular}{lccccccccc
}
\toprule
Method  & \multicolumn{3}{c}{MNIST} & \multicolumn{3}{c}{CIFAR-10} & \multicolumn{3}{c}{Tiny imagenet}\\ 
\cmidrule(lr){2-4} \cmidrule(lr){5-7} \cmidrule(lr){8-10}
 & \begin{tabular}[c]{@{}c@{}}Recon\\ MSE\end{tabular} & \begin{tabular}[c]{@{}c@{}}Emb \\ MSE \end{tabular} & \begin{tabular}[c]{@{}c@{}} Emb \\ Cos \end{tabular} & \begin{tabular}[c]{@{}c@{}}Recon\\ MSE\end{tabular} & \begin{tabular}[c]{@{}c@{}}Emb \\ MSE \end{tabular} & \begin{tabular}[c]{@{}c@{}} Emb \\ Cos\end{tabular} & \begin{tabular}[c]{@{}c@{}}Recon\\ MSE\end{tabular} & \begin{tabular}[c]{@{}c@{}}Emb \\ MSE \end{tabular} & \begin{tabular}[c]{@{}c@{}} Emb \\ Cos \end{tabular}  \\
\midrule
GRNA & $0.5533\pm0.0919$ & {-} & {-} & $0.3287\pm0.0061$& {-} & {-} & $0.3869\pm0.0103$& {-} & {-}\\
GIA  & $0.9125\pm0.0056$ & $ 0.0179\pm0.0018$ & $0.2772\pm0.0431$ & $0.2550\pm0.0004$ & $ 1.4011\pm0.1849$ & $0.3957\pm0.0535$ &$0.3234\pm0.0009$ & $1.8109\pm0.1039$ & $ 0.3936\pm0.0068$\\
AGN & $ 0.4801\pm 0.0027$& {-} & {-} & $0.4413\pm0.0385$ & {-} & {-}& $0.3786\pm0.0190$ & {-} & {-} \\
AGN-SG  & $0.5131\pm0.0014$ & {-} & {-} & $0.5154\pm0.0776$ & {-} & {-}& $1.2398\pm0.0102$& {-} & {-} \\
AGN-GS& $ 0.7073\pm0.0813$ & {-} & {-} & $0.5901\pm0.0620$ & {-} & {-} & $1.9577\pm0.2531$& {-} & {-} \\
PCAT  & $0.2446\pm0.1312$ & $0.0282\pm0.0016$ & $0.3484\pm0.0299$ & $0.3843\pm0.0123$ &$ 2.4672\pm0.3024$ & $0.5544\pm0.0023$ &$0.3063\pm0.0249$ & $2.7853\pm0.0521$ & $0.5526\pm 0.0025$\\
SDAR  & $0.0839\pm0.0751$ & $0.1638\pm0.1850$ & $0.3142\pm0.1699$ & $0.3067\pm0.0618$ & $0.5230\pm0.6965$ & $ 1.2196\pm 0.4831$ & $ 0.1363\pm0.0293$ & $2.1344\pm0.1174$ & $0.4692\pm 0.0281$\\
FSHA & $0.0536\pm0.0118$ & $0.0168\pm0.0100$ & $0.0909\pm0.0157$ & $0.0317\pm0.0065$ & $0.4657\pm0.1286$ & $0.1610\pm0.0285$ & $0.1025\pm0.0084$ & $1.4897\pm0.2210$ & $0.3056\pm0.0203$ \\
FSHA-SG  & $0.5158\pm0.2511$& $0.0621\pm0.0095$ & $0.2956\pm0.0691$ &  $0.1765\pm0.0390$& $2.1634\pm0.1310$ & $0.4282\pm0.0480$& $0.3110\pm0.1202$ & $2.8927\pm 0.3759$ & $0.4904\pm0.0519$\\
FSHA-GS  & $0.2829\pm0.1374$& $0.1446\pm0.0260$ & $0.3077\pm0.0402$& $0.1312\pm0.0150$&$1.5940\pm0.0684$  &  $0.3133\pm0.0230$&$0.4797\pm0.2383$ & $3.0416\pm0.4313$ & $0.5200\pm0.0423$\\
\begin{tabular}[c]{@{}l@{}}\textbf{URVFL} \\ \textbf{(SG/GS)}\end{tabular}  & $0.0132\pm0.0027$ & $0.0045\pm0.0009$& $\textbf{0.0287}\pm0.0029$ & $0.0302\pm0.0011$ & $0.5679\pm0.0834$ & $0.1303\pm0.0131$&$\textbf{0.0699}\pm0.0055$& $1.4316\pm0.6673$&$\textbf{0.2139}\pm0.0195$ \\
\begin{tabular}[c]{@{}l@{}}\textbf{URVFL\_sync} \\ \textbf{(SG/GS)}\end{tabular} & $\textbf{0.0127} \pm 0.0011$& $\textbf{0.0040}\pm0.0006$ & $0.0341\pm0.0022$ & $\textbf{0.0176}\pm0.0139$ & $\textbf{0.2793}\pm0.1868$ &$\textbf{0.0675}\pm0.0549$& $0.0704\pm0.0011$ & $\textbf{1.1441}\pm0.0058$ & $0.2434 \pm 0.0077$\\
\bottomrule
\end{tabular}
\end{adjustbox}
\vspace{-3mm}
\label{tab:main2}
\end{table*}

\begin{table*}
\centering
\caption{Results of data reconstruction measured by PSNR and SSIM on image datasets. Attack methods followed by SG or GS represent implementations under SG or GS detections. \sys and URVFL\_sync achieve the same performance with and without defenses. $\uparrow$ denotes that the larger the value, the better the performance.}

\begin{adjustbox}{width=\textwidth,center}
\begin{tabular}{lccccccc}
\toprule
Method & \multicolumn{2}{c}{MNIST} & \multicolumn{2}{c}{CIFAR-10} & \multicolumn{2}{c}{Tiny imagenet}\\ 
\cmidrule(lr){2-3} \cmidrule(lr){4-5} \cmidrule(lr){6-7}
 & PSNR$\uparrow$ & SSIM $\uparrow$& PSNR$\uparrow$ & SSIM$\uparrow$ & PSNR$\uparrow$ & SSIM$\uparrow$ \\
\midrule
GRNA & 8.8317±0.0005 & 0.1467±0.0001 & 11.3939±0.0074 & 0.0019±0.0001 & 7.3986±0.5618 & 0.0045±0.0001 \\
GIA & 6.4411±0.0006 & 0.0029±0.0001 & 11.9625±0.0001 & 0.0018±0.0001 & 10.9342±0.0002 & 0.0010±0.0000 \\
AGN & 9.2787±0.2651 & 0.1400±0.0033 & 10.4527±0.1293 & 0.0150±0.0001 & 10.2442±0.0487 & 0.0150±0.0001 \\
AGN-SG & 8.9643±0.1067 & 0.0499±0.0015 & 9.0854±1.0272 & 0.0156±0.0001 & 3.4810±0.0000 & 0.0006±0.0256 \\
AGN-GS & 7.8738±0.6595 & 0.0746±0.0027 & 7.7755±0.9967 & 0.0142±0.0001 & 3.7093±0.4056 & 0.0034±0.0001 \\
PCAT & 14.0504±0.4956 & 0.4506±0.0177 & 11.9012±1.6192 & 0.0727±0.0053 & 11.3199±0.1573 & 0.0160±0.0001 \\
SDAR & 17.8807±0.6992 & 0.7093±0.0017 & 14.9388±2.3774 & 0.2525±0.0121 & 14.7862±0.9083 & 0.1318±0.0017 \\
FSHA & 22.7711±2.2248 & 0.8901±0.0006 & 19.7955±2.2895 & 0.5312±0.0298 & 15.9321±0.1248 & 0.2205±0.0003 \\
FSHA-SG & 10.4247±1.6129 & 0.1928±0.0176 & 13.2670±0.5318 & 0.1294±0.0006 & 12.6357±0.7394 & 0.0395±0.0002 \\
FSHA-GS & 11.1577±4.5247 & 0.2151±0.0137 & 15.1751±0.5590 & 0.3083±0.0053 & 12.4729±1.9752 & 0.0577±0.0001 \\
URVFL(GS/SG) & 25.1349±1.2354 & \textbf{0.9385}±0.0002 & 21.0666±0.0165 & 0.5927±0.0002 & \textbf{17.6107}±0.0992 & 0.3023±0.0001 \\
URVFL\_sync(GS/SG) & \textbf{25.2919}±0.1369 & 0.9324±0.0002 & \textbf{26.2360}±0.0030 & \textbf{0.8125}±0.0001 & 17.5381±0.0029 & \textbf{0.3053}±0.0001 \\
\bottomrule
\end{tabular}
\end{adjustbox}
\label{tab:main3}
\end{table*}

\subsection{Baselines and detections}
To evaluate the efficacy of our methods, we compare with three prominent data reconstruction strategies in VFL, as well as adaptations of attacks from SL. For HBC settings in VFL, we select GRNA and GIA, and for the malicious setting, we consider AGN. Additionally, we adapt three SL strategies for VFL: HBC attacks PCAT and SDAR, and the malicious attack FSHA. 

\noindent \textbf{GRNA:} The adversary uses a generator model to create synthetic features. These synthetic features are then used to query the target model to update the generator, which minimizes the distance between the predictions made from synthetic features and those from real features. 
    
\noindent \textbf{GIA:} This approach involves training a shadow model to emulate the target model's behavior.  Then the adversary feeds random noise into the trained shadow model and optimizes noise to reduce the distance between the noise embeddings and real data embeddings. 
 
 \noindent\textbf{AGN:} AGN discards the top model and instead uses a generator to produce synthetic data from the adversary's features and aggregated embeddings. Concurrently, a discriminator works to differentiate between the generator's output and genuine data samples.
 
  \noindent\textbf{PCAT:} PCAT uses the top model to guide a shadow model, which simulates the behavior of the target model. A decoder is trained on the shadow model to reconstruct the target features.

 \noindent\textbf{SDAR:} SDAR builds on PCAT by adding two discriminators that evaluate both the embeddings from the shadow model and the reconstructed features. 

\noindent \textbf{FSHA:} In FSHA, an embedding discriminator replaces the top model. Besides, a shadow model and a decoder are trained to reconstruct target features. 

We also implement SOTA detection methods, SplitGuard (SG) and Gradient Scrutinizer (GS) on malicious attacks, to assess reconstruction performance under these defenses.

 \noindent\textbf{SplitGuard (SG):} Analyzes gradients from fake and regular batches to compute a detection score after each round of fake batches. Following the original method, the target client calculates an average score from the last 10 records; if it falls below a threshold of 0.9, an attack is detected, and updates to the bottom model are halted.

\noindent\textbf{Gradient Scrutinizer (GS):} Evaluates the detection score of received gradients in each round to identify potential attacks. Due to instability in detection scores leading to frequent misjudgments, we adopt an average score strategy to improve reliability and set the threshold to 0.8 in our experiments. Similarly, when the average detection score falls below the detection threshold, an attack is detected and the bottom model update stop.

In our experimental results (Table~\ref{tab:main1}, \ref{tab:main2}, and \ref{tab:main3}), we append the GS or SG to the names of baseline methods to indicate their evaluation under detection mechanisms, SG or GS. 

\subsection{Data reconstruction results}


\noindent\textbf{Overview:}
As shown in Table~\ref{tab:main1}, \ref{tab:main2}, and \ref{tab:main3}, for all datasets, our proposed \sys and URVFL\_sync consistently outperform all other approaches in achieving the lowest reconstruction errors and embedding cosine distance. Our methods also achieve the best reconstruction performance measured by PSNR and SSIM compared with other baselines in image datasets. This underscores the effectiveness of our techniques and the advantage of embedding distribution transfer using DAC.
For the Tiny imagenet, URVFL\_sync achieves a reconstruction error of 0.0699, PSNR 17.5381, SSIM 0.3053, and cosine distance of 0.2139, significantly outperforming FSHA, which has a reconstruction error of 0.1025, PSNR 15.9321, SSIM 0.2205, and cosine distance of 0.3056.

Our methods demonstrate robustness against detection mechanisms, maintaining consistent performance and circumventing both detection methods malicious attacks, i.e., URVFL and URVFL\_sync is not detected by SG and GS. In sharp contrast, the performance of AGN and FSHA become worse when faced with the detections, and even completely failed under some circumstances. 
Specifically, AGN's reconstruction error increases dramatically from 0.3786 to 1.9577, and FSHA's from 0.1025 to 0.4797 for the Tiny imagenet dataset. 
This demonstrates the stealthiness and adaptability of \sys and URVFL\_sync compared to other methods.

An observation from the results is that a smaller embedding MSE distance does not guarantee a more accurate reconstruction or a more similar embedding distribution, as evidenced by GIA in the Credit dataset and PCAT in the RT\_IOT2022 dataset, which has the smallest embedding MSE distance but significantly higher cosine embedding distance and reconstruction error compared to our methods. While minimizing the embedding MSE distance helps reconstruction, it does not guarantee a more similar embedding distribution, as metrics like cosine distance better capture the similarity between distributions. Generally, the decoder is trained to reconstruct data from the embedding distribution. Therefore, even if our methods have a higher embedding MSE distance, as long as the target model generates a similar embedding distribution, the decoder can still work effectively.

Additionally, a smaller embedding cosine distance does not necessarily correspond to a lower reconstruction error in \sys and URVFL\_sync.
For example, for the MNIST and RT\_IOT2022 dataset, while URVFL\_sync achieves a lower reconstruction error than \sys, it exhibits a slightly higher embedding cosine distance. This can be attributed to the synchronous training of all model components in URVFL\_sync, which promotes dynamic learning and results in model diversity. Despite the higher average embedding  distance, the decoder in URVFL\_sync is more effective at reconstructing features, effectively compensating for the increased cosine distance.

\begin{figure*}[ht]
\centering

\begin{subfigure}{0.19\textwidth}
    \includegraphics[width=\linewidth]{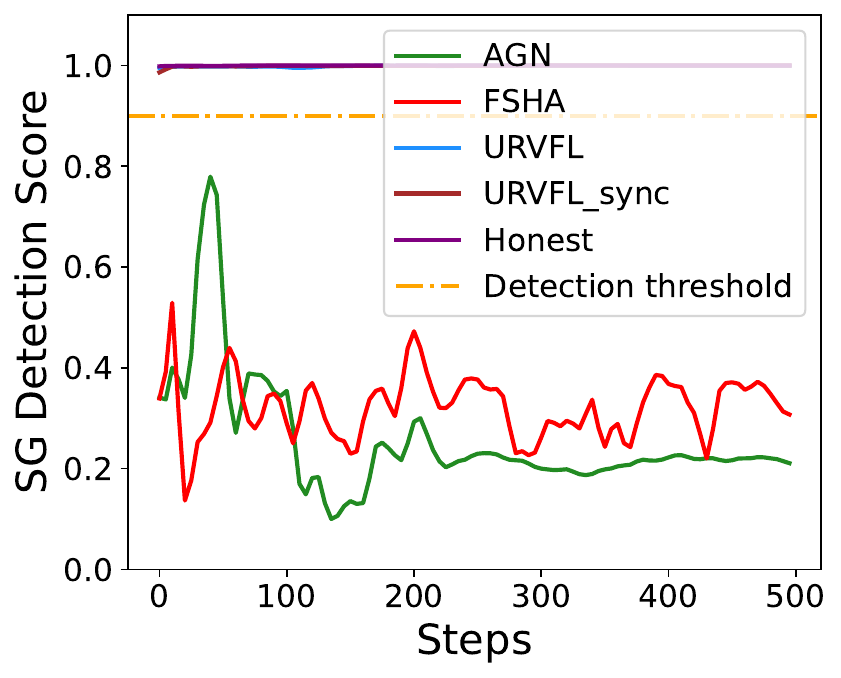}
    \caption{SG scores in MNIST.}
    \label{fig:sub1a}
\end{subfigure}
\hspace{0pt}
\begin{subfigure}{0.19\textwidth}
    \includegraphics[width=\linewidth]{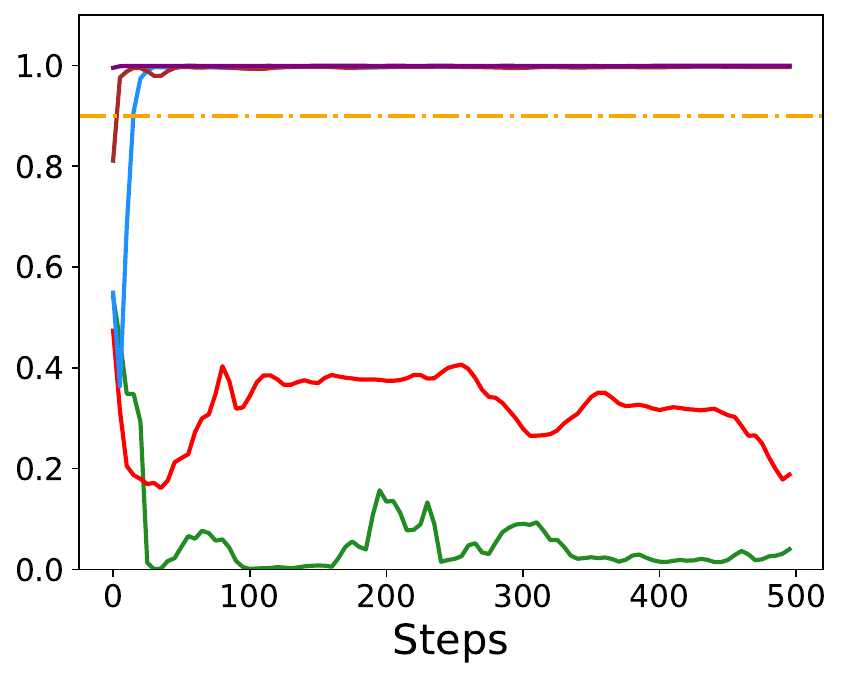}
    \caption{SG scores in CIFAR-10.}
    \label{fig:sub1b}
\end{subfigure}
\hspace{0pt}
\begin{subfigure}{0.19\textwidth}
    \includegraphics[width=\linewidth]{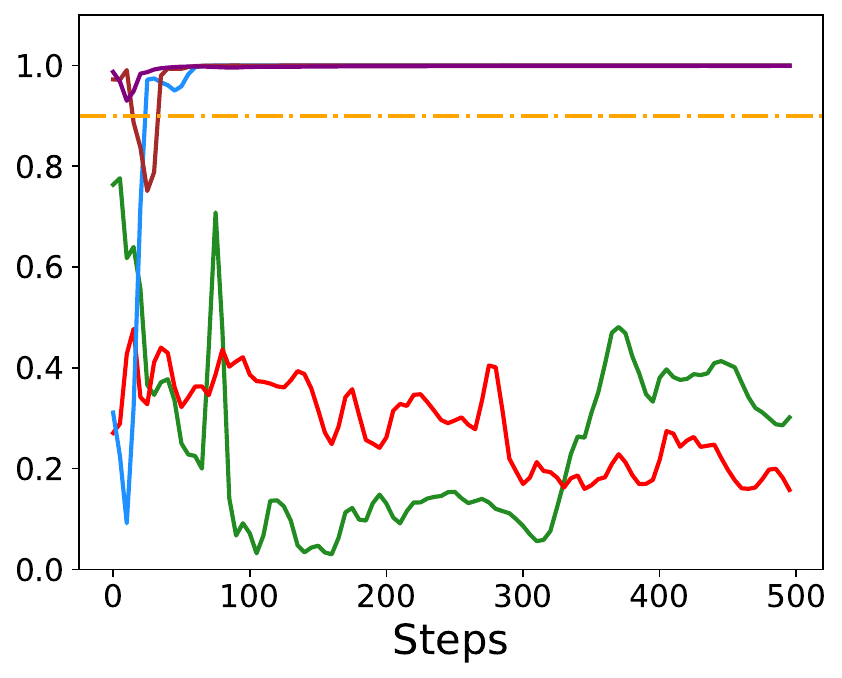}
    \caption{SG scores in Tiny.}
    \label{fig:sub1d}
\end{subfigure}
\hspace{0pt}
\begin{subfigure}{0.19\textwidth}
    \includegraphics[width=\linewidth]{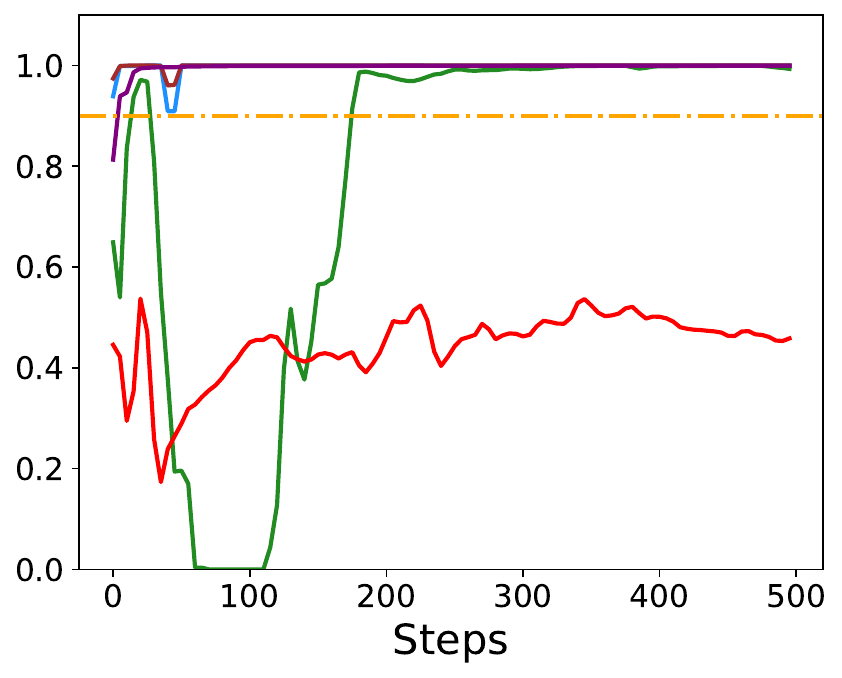}
    \caption{SG scores in IOT.}
    \label{fig:sub1c}
\end{subfigure}
\hspace{0pt}
\begin{subfigure}{0.19\textwidth}
    \includegraphics[width=\linewidth]{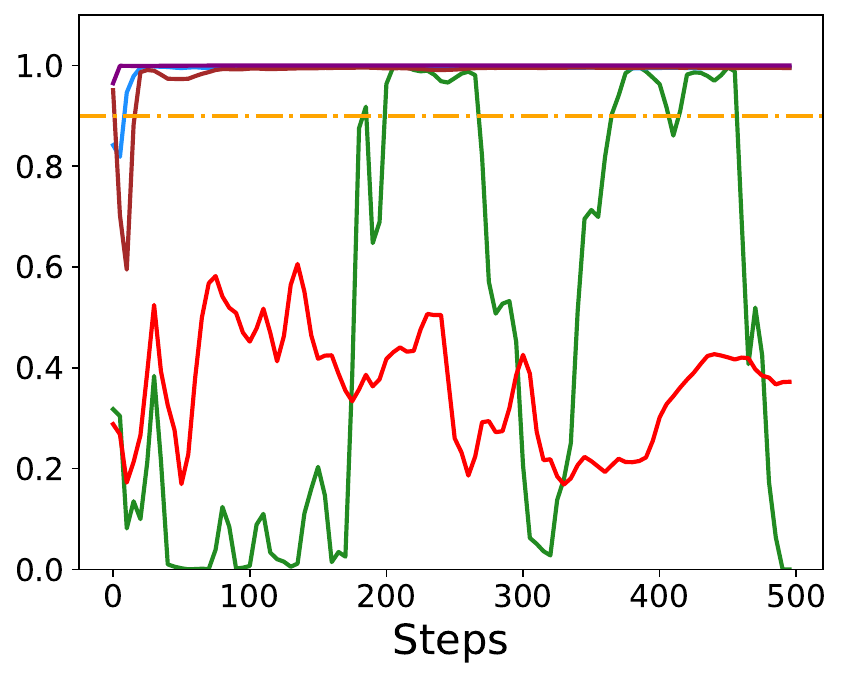}
    \caption{SG scores in Credit.}
    \label{fig:sub1d}
\end{subfigure}

\begin{subfigure}{0.19\textwidth}
    \includegraphics[width=\linewidth]{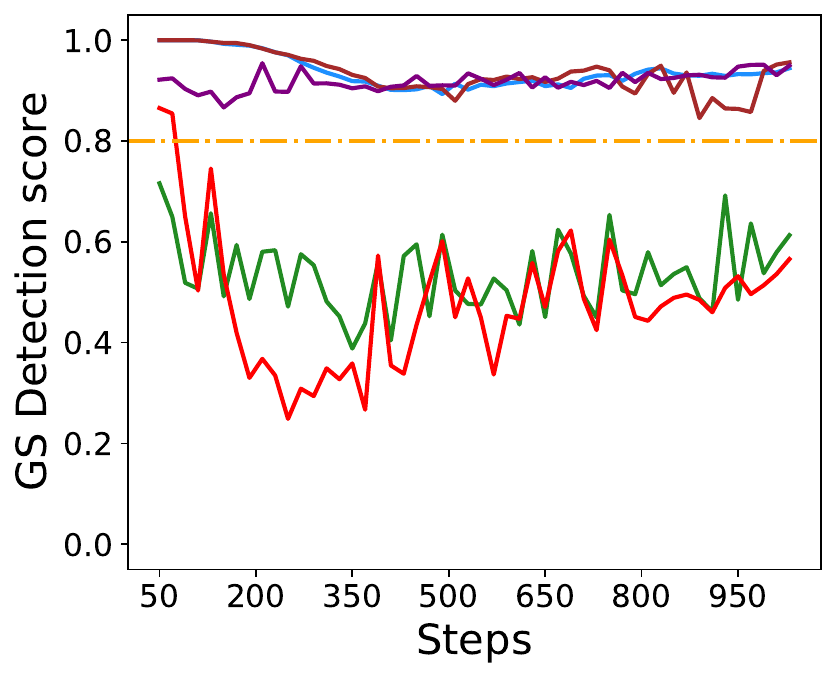}
    \caption{GS scores in MNIST.}
    \label{fig:sub2a}
\end{subfigure}
\hspace{0pt}
\begin{subfigure}{0.19\textwidth}
    \includegraphics[width=\linewidth]{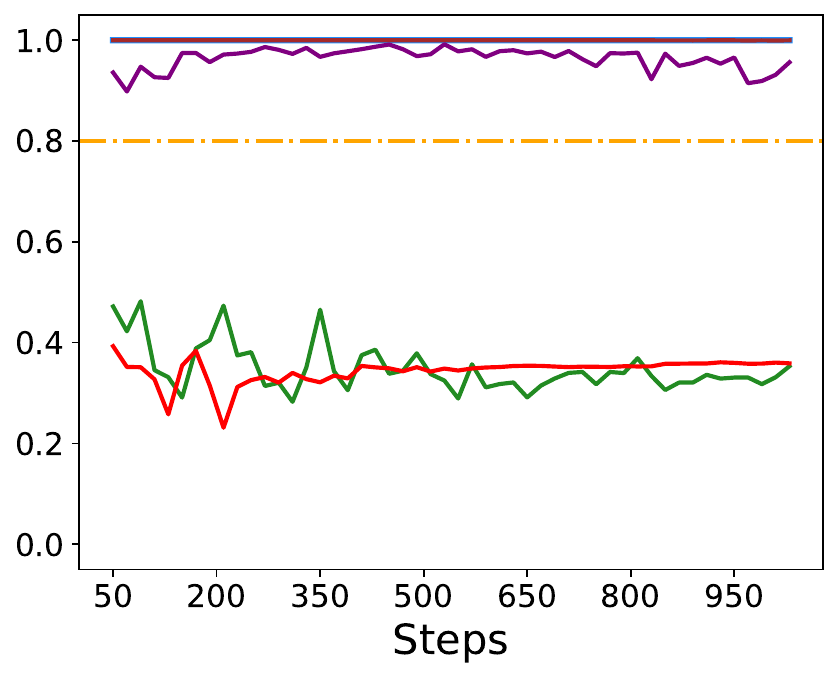}
    \caption{GS scores in CIFAR-10.}
    \label{fig:sub2b}
\end{subfigure}
\hspace{0pt}
\begin{subfigure}{0.19\textwidth}
    \includegraphics[width=\linewidth]{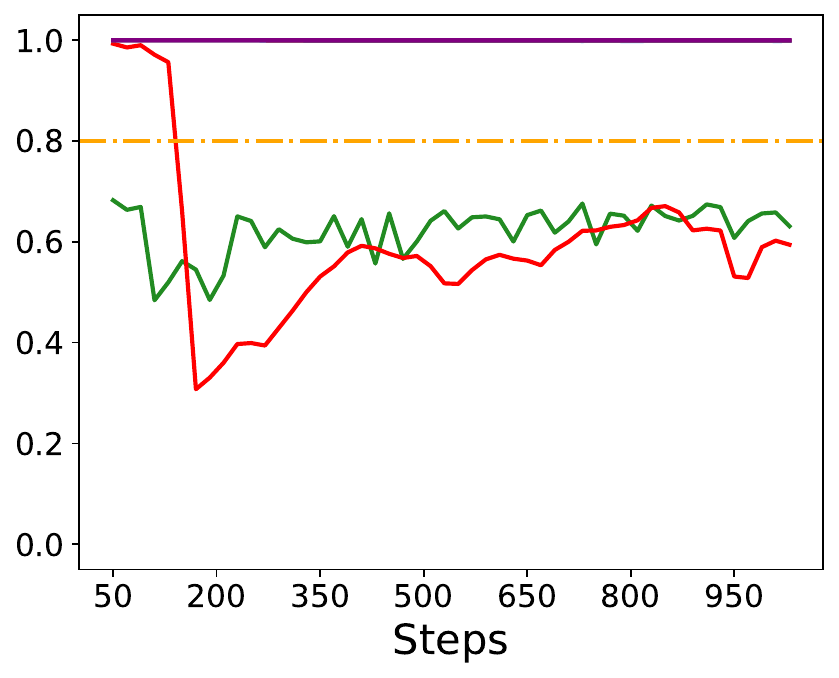}
    \caption{GS scores in Tiny.}
    \label{fig:sub1d}
\end{subfigure}
\hspace{0pt}
\begin{subfigure}{0.19\textwidth}
    \includegraphics[width=\linewidth]{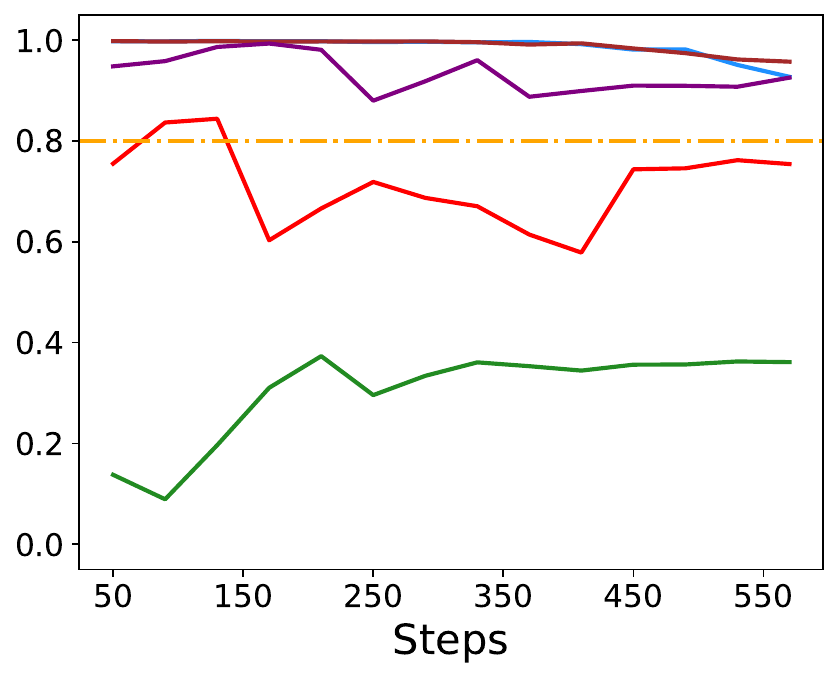}
    \caption{GS scores in IOT.}
    \label{fig:sub2c}
\end{subfigure}
\hspace{0pt}
\begin{subfigure}{0.188\textwidth}
    \includegraphics[width=\linewidth]{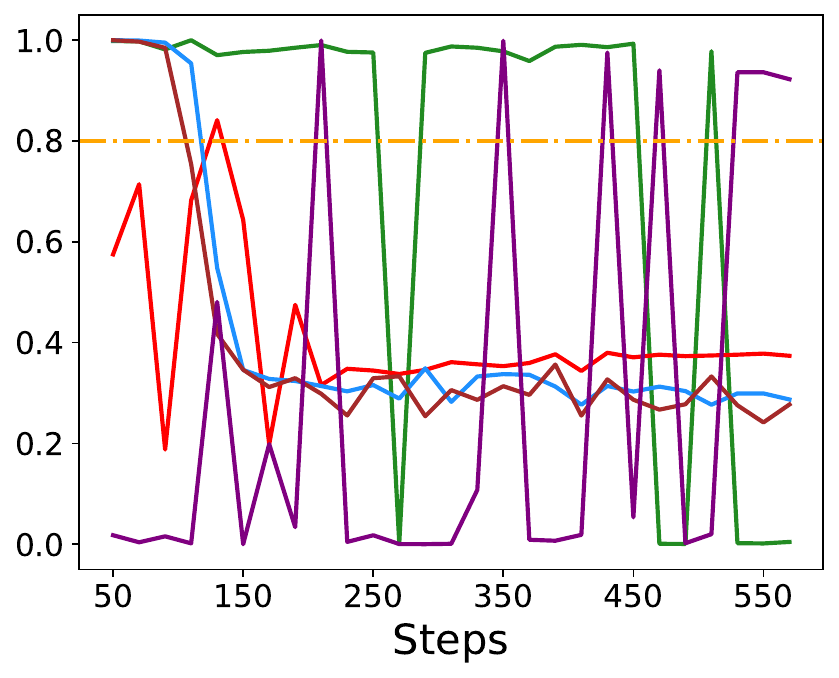}
    \caption{GS scores in Credit.}
    \label{fig:sub2d}
\end{subfigure}
\hspace{0pt}
\vspace{-3mm}
\caption{Detection scores of SG and GS on malicious attacks and the honest training across four datasets. We abbreviate Tiny imagenet dataset to Tiny and RT\_IOT2022 dataset to IOT.}
\vspace{-4mm}
\label{fig:defense}
\end{figure*}

We do not record AGN-GS and FSHA-GS in the Credit dataset. It is because GS doesn't work in the Credit dataset, which can also be observed in Figure~\ref{fig:sub2d}, caused by the limited data features (23 features) and imbalanced label distribution (77\% label 0 and 23\% label 1).

\noindent\textbf{Visualization of the reconstruction:} Table~\ref{tab:main_visual} visualizes the reconstruction performance of the proposed \sys, alongside the malicious attack baselines, AGN and FSHA, for the CIFAR-10 dataset. We also report the visualization result of Tiny imagenet in Table~\ref{tab:visual_tiny} in Appendix~\ref{tinyrecon}. The results clearly demonstrate that \sys produces more accurate and distinct reconstructions compared to the baselines. While AGN and FSHA yield coarser results, \sys maintains high-quality reconstruction even under SG detection. In contrast, both FSHA and AGN struggle to reconstruct informative pixels effectively when faced with detection mechanism SG. 

\begin{table*}[ht]
\centering
\caption{Visualization of reconstructed features on the CIFAR-10 dataset with and without the detection SG. Except for the original image, the right side of each image is the reconstructed image. The last column is the average reconstruction error. }
\includegraphics[width=1\linewidth]{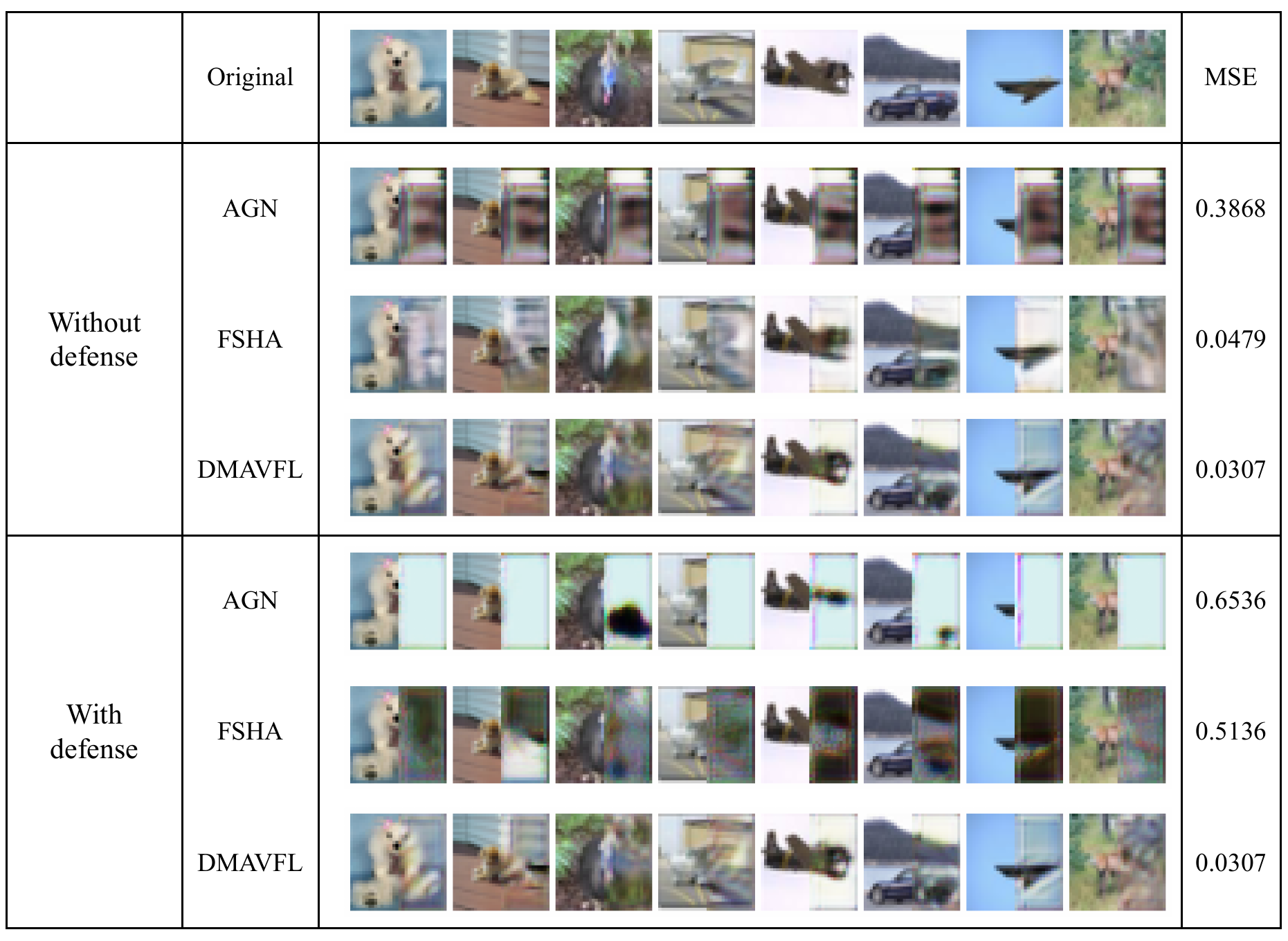}
\label{tab:main_visual}
\end{table*}

\noindent\textbf{SG and GS detection results: }Figure~\ref{fig:defense} illustrates the detection scores for \sys and URVFL\_sync, which consistently exceed the detection thresholds, aligning indistinguishably with honest training scenarios. Although in the CIFAR-10 and Tiny imagenet dataset, \sys's detection score is lower than the threshold at the beginning, its average score remains above the threshold, ensuring stealthiness against target clients. In contrast, malicious attack baselines, AGN and FSHA, consistently maintain scores below the detection thresholds of SG and GS in the MNIST and CIFAR-10 datasets, rendering them detectable.

In the RT\_IOT2022 and Credit datasets under defense SG, AGN sporadically surpasses the threshold, yet its average detection score falls below the threshold, leading passive clients to detect an attack promptly. FSHA’s SG scores remain below the threshold throughout, making it easily detectable. Under GS, FSHA exceeds the threshold at the beginning but is ultimately detected in the RT\_IOT2022 dataset. In the Credit dataset, as previously noted, GS is ineffective due to imbalanced labels and limited features.

\begin{figure}[htbp]

\begin{minipage}{0.24\textwidth}
\centering
\includegraphics[width=\textwidth]{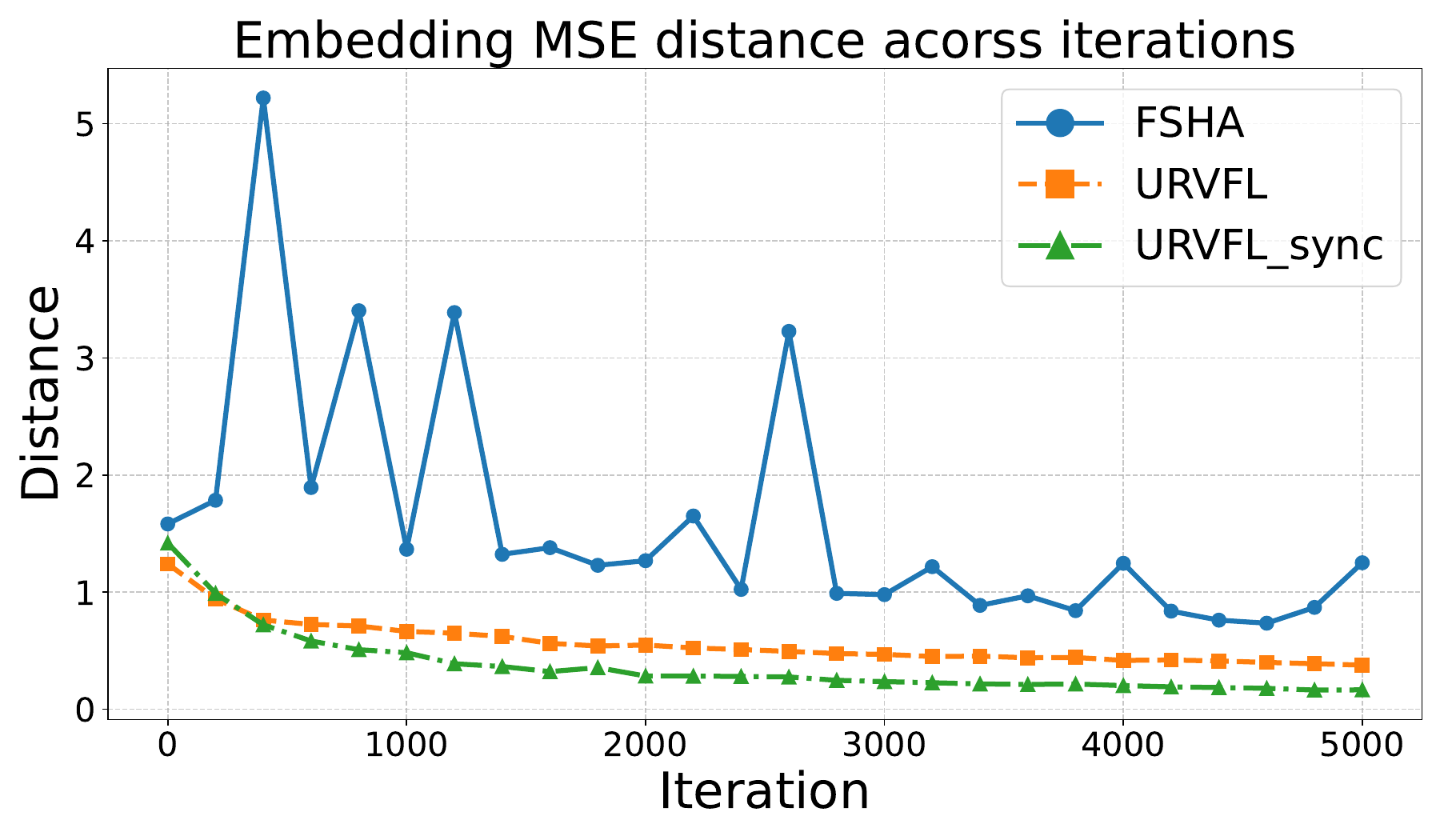}
\label{fig:recon_feature}
\end{minipage}
\vspace{-3mm}
\hfill
\begin{minipage}{0.24\textwidth}
\centering
\includegraphics[width=\textwidth]{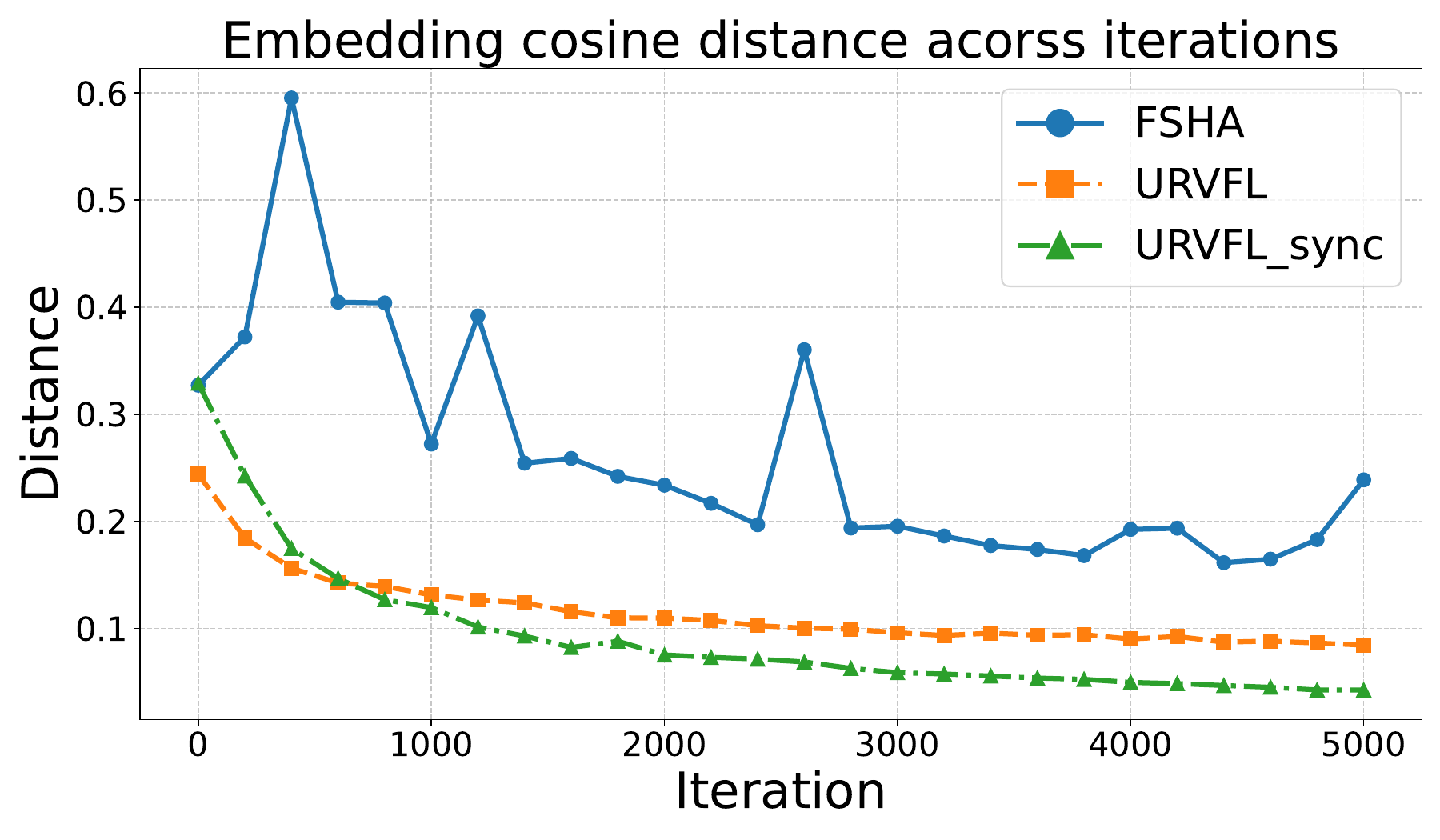}
\label{fig:reconstruction_error}
\end{minipage}
\vspace{-3mm}
\caption{Embedding MSE and cosine distances across attack iterations using FSHA, URVFL, and URVFL\_sync.}
\vspace{-3mm}
\label{main: distance}
\end{figure}

\noindent\textbf{Embedding distances across attack iterations}: Figure~\ref{main: distance} visualizes the changes in embedding distance during the attack for our methods compared to FSHA on the CIFAR-10 dataset. The figure clearly demonstrates that our methods steadily reduce the embedding distance throughout the attack, resulting in lower overall embedding distances and effectively aligning the embedding distribution with the target model. In contrast, FSHA exhibits less stability, despite also showing a decreasing trend in embedding distance.

Furthermore, we observe that the SG score for a single iteration of our methods falls below the threshold within the first 10 iterations in Fig.~\ref{fig:defense}, whereas FSHA remains below the threshold throughout the entire training process, leading to the detection of the attack. This difference may be attributed to the more drastic changes in embedding distance during our training process. As illustrated in Figure~\ref{main: distance}, our methods cause rapid changes in embedding distance initially, which then slow down, while FSHA maintains a more consistent fluctuations in embedding distance throughout the attack.

%% file: ablationstudy.tex
\subsection{Ablation Study}
This section explores the robustness of \sys and URVFL\_sync (abbreviated as sync) by evaluating their performance under various parameter settings. These settings include changes in the adversary’s feature size, use of heterogeneous encoder models, variations in auxiliary dataset size, and differences in the number of target clients. Except for the variables specifically altered for this study, the model setups and implementation environment remain consistent with those described in the main experiments.

\subsubsection{Effect of adversary's feature size}

\begin{figure}[htbp]
    \begin{minipage}{0.24\textwidth}
        \centering
        \includegraphics[width=\textwidth]{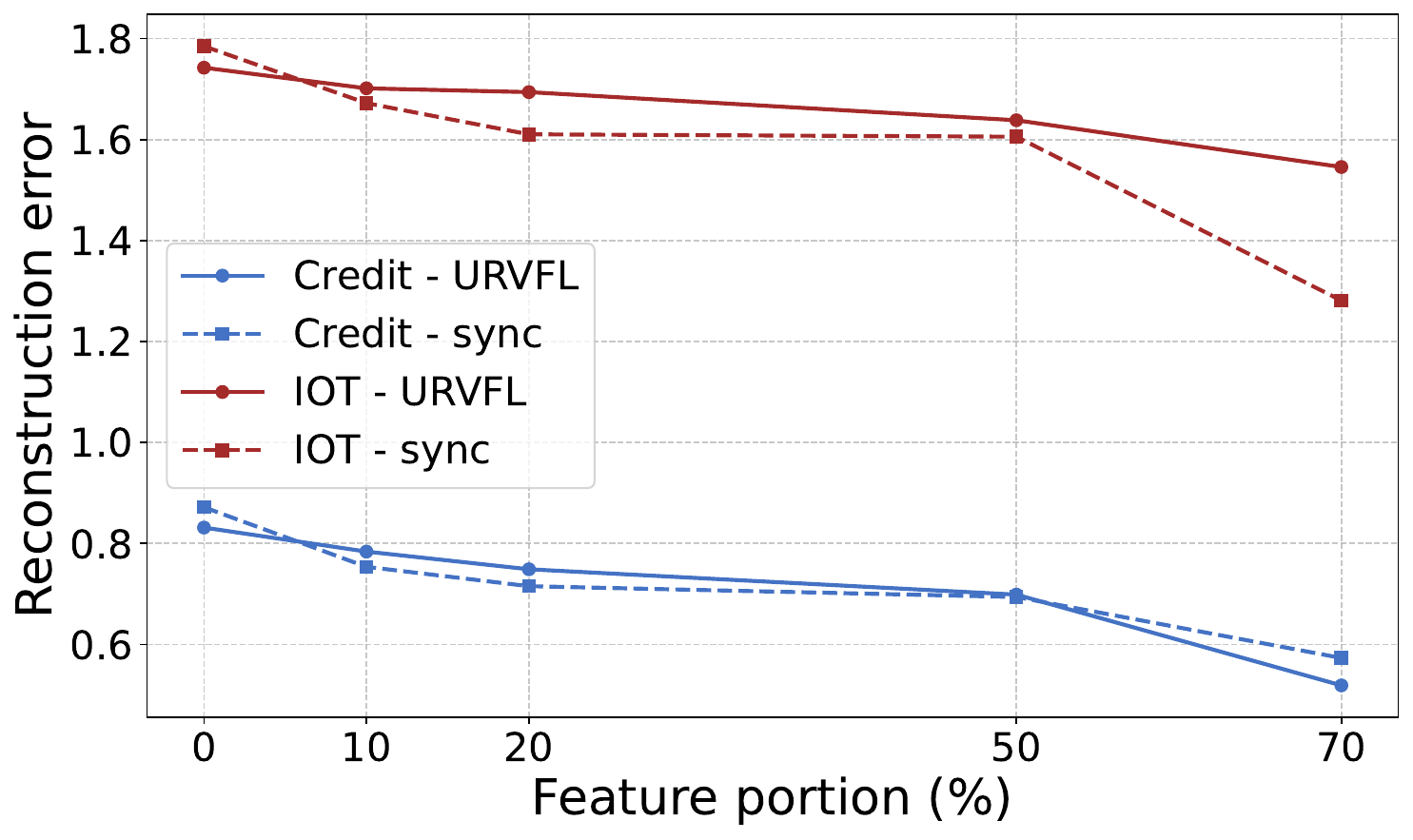}
        \caption{Reconstruction error under various adversary's feature portions.}
        \label{fig:recon_feature}
    \end{minipage}
    \hfill
    \begin{minipage}{0.24\textwidth}
        \centering
        \includegraphics[width=\textwidth]{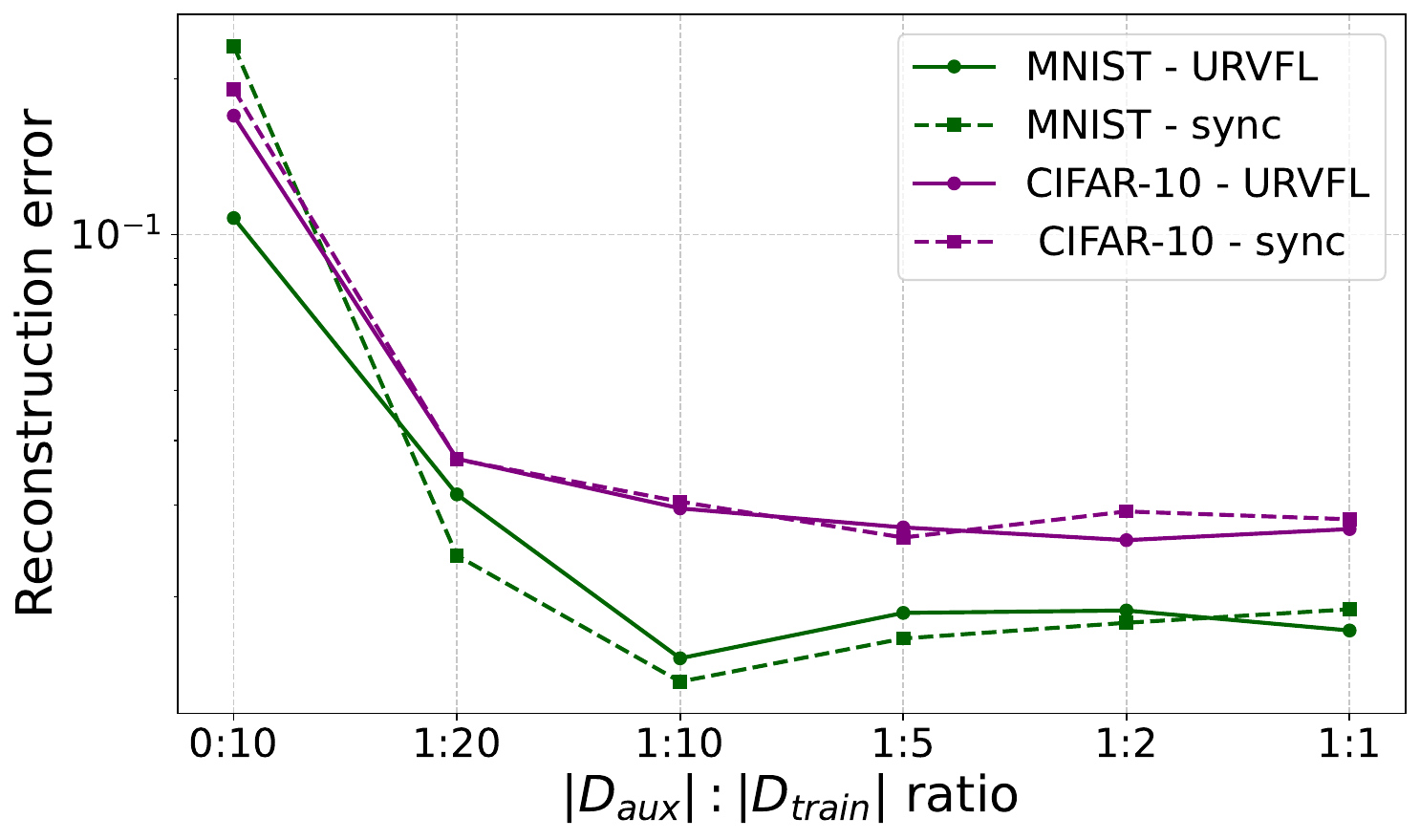}
        \caption{Reconstruction error for different $|D_{aux}|$ relative to $|D_{train}|.$}
        \label{fig:reconstruction_error}
    \end{minipage}
    \vspace{-3mm}
\end{figure}

To explore how the size of the feature set held by the adversary affects the reconstruction performance of \sys and sync, we conduct experiments targeting the reconstruction of 30\% of the features while varying the proportion of features held by the adversary from 0\% to 70\%. At 0\% feature possession, the scenario reduces to split learning where no features or bottom model are held by the adversary. These experiments are performed on the Credit and RT\_IOT2022 (abbreviated as IOT) datasets, with the results summarized in Figure~\ref{fig:recon_feature}.

Consistent with our expectations, increasing the proportion of features held by the adversary leads to less reconstruction error. As illustrated in Figure~\ref{fig:recon_feature}, the reconstruction error in the Credit dataset using \sys decreases from 0.8315 to 0.5189 as the feature possession increases from 0\% to 70\%. This trend is observed for both datasets and methods. 


\subsubsection{Effect of encoder's complexity}
In our main experiments, the encoder mirrors the target model's structure. However, when the adversary lacks precise knowledge about the target model's configuration, discrepancies between the encoder and the target model's structures can affect the transfer of embedding distributions and, consequently, the overall reconstruction performance. To evaluate the impact of varying encoder complexities, we conduct experiments on the CIFAR-10 dataset, where the target model comprises \textbf{3 residual blocks}.
We test the following encoder configurations to assess their effect on performance. \textbf{MLP Model}: Comprises 3 linear layers, with input and output dimensions matching those of the target model; \textbf{1-Res Model}: Utilizes a single residual block.
\textbf{5-Res Model}: Extends to 5 residual blocks, increasing complexity.
We test the reconstruction error and measure both the embeddings MSE and cosine distances for each encoder configuration. 
\begin{table}[htbp]
  \centering
  \caption{Reconstruction error and average embeddings distance using different encoder's model.}

  \footnotesize
  \setlength\tabcolsep{8pt}
  \label{tab:abl_model}
  \begin{tabular}{@{}llcc@{}}
    \toprule
    Encoder & Metric & \sys & sync \\
    \midrule
    \multirow{3}{*}{MLP} & Recon MSE & 0.2545 & 0.2362 \\
           & Emb MSE dis & 7.7767 & 3.9726 \\
           & Emb Cos dis & 0.6337 & 0.6081 \\
    \addlinespace
    \multirow{3}{*}{1-Res} & Recon MSE & 0.0561 & 0.0434 \\
         & Emb MSE dis & 0.3175 & 0.3059 \\
         & Emb Cos dis & 0.1632 & 0.1450 \\
    \addlinespace
    \multirow{3}{*}{\begin{tabular}[c]{@{}c@{}}3-Res\\ (Same)\end{tabular}} & Recon MSE &\textbf{ 0.0302} & \textbf{0.0176} \\
            & Emb MSE dis & 0.5679 & 0.2793 \\
            & Emb Cos dis & 0.1303 & 0.0675 \\
    \addlinespace
    \multirow{3}{*}{5-Res} & Recon MSE & 0.0313 & 0.0366 \\
                 & Emb MSE dis & 0.8771 & 0.7071 \\
                 & Emb Cos dis & 0.1081 & 0.1112 \\
    \bottomrule
  \end{tabular}
  \vspace{-3mm}
\end{table}

Table~\ref{tab:abl_model} clearly illustrates that the adversary achieves the lowest reconstruction error and the smallest embedding cosine distance when using an encoder the same as the target model, as observed for both \sys and sync. 
This optimal alignment underscores the importance of model congruence in enhancing the effectiveness of embedding distribution transfer and data reconstruction. 
Conversely, when the encoder structure deviates from that of the target model, specifically when switching from a CNN to an MLP, the reconstruction performance significantly deteriorates, with reconstruction error increasing by an order of magnitude (from 0.0302 to 0.2545). 
This dramatic drop highlights the challenges of using less compatible models for data reconstruction attack. 
Moreover, the results indicate that a complex encoder, featuring a higher number of residual blocks, outperforms a simpler encoder configuration. The complex encoder not only enhances the decoder's ability to reconstruct the data accurately but also learns a more informative embedding. This suggests that increasing the encoder's complexity, when the target model is unknown, can lead to better performance in both embedding distribution transfer and data reconstruction.

\subsubsection{Effect of auxiliary dataset}
In the main experiments, we set the size of the adversary's auxiliary dataset, $|D_{aux}|$, relative to the training set size, $|D_{train}|$, at a ratio of $1:10$. To explore the impact of varying auxiliary dataset sizes, we adjust the ratio of $|D_{aux}| : |D_{train}|$ from $0 : 10$ to $1:1$ in the MNIST and CIFAR10 dataset. Specifically, 
at a ratio of 0:10, the auxiliary dataset of MNIST and CIFAR-10 comprise the FashionMNIST~\cite{xiao2017fashion} and the downsampled Tiny imagenet, respectively, introducing a distinct distribution from the training set. This setup aims to test the adaptability of the reconstruction process to changes in the auxiliary dataset’s distribution.

From Figure~\ref{fig:reconstruction_error} we can see that significant data leakage can occur, even in the absence of auxiliary data with the same distribution of the training set (i.e., with out-of-distribution auxiliary data). This observation is highlighted by a reconstruction error of 0.1076$\sim$0.2311, when the adversary uses FashionMNIST data to reconstruct images from MNIST, and 0.1696$\sim$0.1906 when the adversary uses Tiny imagenet data to reconstruct images from CIFAR-10. 
We also observe that for both \sys and sync, the reconstruction error decreases quickly as $|D_{aux}|$ becomes relatively larger. However, after the ratio of $|D_{aux}| : |D_{train}|$ reaches 1:10, increasing auxiliary dataset size results in minor improvements in reconstruction error. This suggests that there is a threshold beyond which additional auxiliary data does not substantially improve the model’s ability to reconstruct target features, pointing to a diminishing return on the utility of expanding the auxiliary set.


\subsubsection{Effect of the number of target clients}

Our proposed methods, \sys and sync, are designed to handle scenarios involving multiple target clients. 
To assess the impact of varying the number of target clients on data reconstruction performance, we conduct experiments on the IOT dataset with the number of target clients ranging from 1 to 5. Consistent with the setup in the main experiments, 50\% of the features are equally distributed among the target clients. 

\begin{table}[htbp]
  \centering
  \caption{Reconstruction errors for different numbers of target clients.}
  \footnotesize
  \label{tab:abla_multi}
  \begin{tabular}{@{}lccccc@{}}
    \toprule
   \# of target clients &  1 &  2 & 3 & 4 &  5 \\
    \midrule
    \sys      & {\bf 1.3821} & 1.5801 & 1.5986 & 1.6718 & 1.9820 \\
   sync & {\bf 1.3277} & 1.6447 & 1.7597 & 1.8660 & 2.2635 \\
    \bottomrule
  \end{tabular}
\end{table}

The results presented in Table~\ref{tab:abla_multi} indicate that an increase in the number of target clients, each sharing a smaller portion of the features, leads to diminished reconstruction performance. Specifically, the reconstruction error increases from 1.3821 with a single target client to 1.9820 when there are five target clients. This degradation in performance can be attributed to the distribution of features among more clients, each employing a separate bottom model to learn their respective features. This setup tends to cause the loss of crucial feature correlation information during the learning process. Moreover, as the encoder is trained on the complete set of features and is hence able to capture more feature correlations, the increased disparity between the encoder and the target models' understanding of the feature correlation further exacerbates the challenge of accurate data reconstruction.

\section{Evaluations on defenses}
While \sys is designed and demonstrated to penetrate detection-based methods of SplitGuard and Gradient Scrutinizer, we also evaluate its effectiveness against general defensive methods for privacy protection in VFL and SL. \emph{Note that unlike SplitGuard and Gradient Scrutinizer, these methods essentially introduce perturbations to the communicated embeddings or gradients, severely degrading the performance of honest training.} Specifically, we consider three such defenses: Nopeek~\cite{vepakomma2020nopeek}, obfuscation with random noise~\cite{he2020attacking}, and differential privacy (DP)~\cite{dwork2006differential, abadi2016deep,wei2020federated}. 


\noindent \textbf{Nopeek Defense.} Nopeek is a method for minimizing distance correlation, widely implemented in both VFL and SL to prevent privacy leakage. It operates by minimizing the correlation between embeddings and the original features to reduce information leakage. During VFL training, passive clients incorporate the Nopeek loss into their training loss, defined as:
\begin{equation}
\begin{aligned}
    \mathcal{L}_{N} &= \frac{1}{|\mathcal{B}_{train}|}\sum_{j\in \mathcal{B}_{train}} (\alpha\cdot DCOR(\boldsymbol{x}_{j,n}, f_n(\boldsymbol{x}_{j,n})) \\ &+ (1-\alpha)\cdot TASK(y_j,f_n(\boldsymbol{x}_{j,n} )),
    \end{aligned}
\end{equation}
where $DCOR$ represents the distance correlation metric, $TASK$ denotes the VFL training loss, and $\alpha$ is a hyper-parameter balancing training and defense.

We challenge Nopeek by deploying \sys and sync against it with varying $\alpha$ values from 0.1 to 1.0 on the MNIST dataset. The results in Table~\ref{tab:nopeek} show that 
when $\alpha < 0.9$, Nopeek fails to thwart our attacks. 
As $\alpha$ approaches 0.9 and beyond, Nopeek begins to impede the attack effectiveness. Notably, even at $\alpha = 0.95$, \sys still achieves a small reconstruction error of 0.0280. Setting $\alpha = 1.0$ effectively defends the attack. However, this completely replaces the original VFL task with minimizing the $DCOR$ loss, which results in an accuracy of merely 69.03\%.
\begin{table}[htbp]
  \centering
  \caption{Reconstruction error against Nopeek.} 
  \scriptsize
  \setlength\tabcolsep{2pt}
  \label{tab:nopeek}
  \begin{tabular}{@{}lccccccccc@{}}
    \toprule
    $\alpha$ & 0 & 0.2 & 0.4 & 0.6 & 0.8& 0.9&0.95&1.0 \\
    \midrule
    \sys & 0.0132 & 0.0151 & 0.0122 & 0.0138 & 0.0116 &0.0171&0.0280&0.4759\\
    sync   & 0.0127 & 0.0143 & 0.0122 & 0.0113 & 0.0135&0.0156&0.0403&0.4462 \\
    Honest Acc(\%) &99.11 & 97.82& 98.28 & 98.04 & 97.90& 95.03& 94.72 & 69.03\\
    \bottomrule
  \end{tabular}
\end{table}


\noindent \textbf{Obfuscation with random noise.} Perturbation methods, such as adding noise to transmitted embeddings, are commonly employed in ML to safeguard privacy. Following the approach outlined in \cite{he2020attacking}, we introduce noise with standard deviations $\sigma$ ranging from 0.2 to 0.8 into the embeddings sent by target clients. This defense is tested against \sys and sync on the MNIST dataset, and the results are reported in Table~\ref{tab:obfuscation}. The results suggest that higher noise levels more effectively impede the attacks. This defense is more effective for sync, as its DAC learns a dynamic distribution of the encoding embedding during training, and is more susceptible to fluctuations caused by noise. Nonetheless, our attacks manage to reconstruct informative features even under significant noise, for instance, \sys achieving a reconstruction error of 0.0472 at $\sigma = 0.7$. It is also important to note that adding larger noise adversely affects the honest training accuracy.
\begin{table}[htbp]
  \centering
  \caption{Reconstruction error against obfuscation.}
  \footnotesize
  \label{tab:obfuscation}
  \begin{tabular}{@{}lcccccc@{}}
    \toprule
    $\sigma$ & 0 & 0.1 & 0.3 & 0.5 & 0.7 \\
    \midrule
    \sys & 0.0132 & 0.0214 & 0.0252 & 0.0351 & 0.0472 \\
    sync   & 0.0127 & 0.0161 & 0.0301 & 0.0497 & 0.0752 \\
    Honest Acc(\%) & 99.11& 98.34 & 97.65 & 94.84& 94.00\\
    \bottomrule
  \end{tabular}
  \vspace{-2mm}
\end{table}

\noindent\textbf{Differential privacy (DP).} DP provides a mathematical framework for quantifying privacy guarantees for ML models. In our attacks, as the gradients returned from the active client may be maliciously generated, we propose to apply $\epsilon$-DP mechanism, which involves adding Laplacian noise to received gradients, to disrupt the malicious training.
We test varying levels of privacy by adjusting the parameter $\epsilon$ from 0.1 to 10, where a smaller $\epsilon$ implies stronger privacy due to higher noise levels. The average reconstruction errors of \sys and sync on MNIST are shown in Table~\ref{tab:dp}, indicating that DP is most effective at an $\epsilon $ of 0.1, successfully impeding our proposed attacks. However, as $\epsilon$ increases, its defensive effectiveness decreases. At $\epsilon = 10$, the noise assists the attack, potentially by making the model updates more robust, thereby facilitating the embedding distribution transfer. It's important to note that while DP helps to protect privacy, it also hurts the accuracy of honest training.
\begin{table}[htbp]
  \centering
  \caption{Reconstruction error against DP. 
  $\epsilon = +\infty$ corresponds to the scenario without DP.}
  \scriptsize
  \setlength\tabcolsep{3.5pt}
  \label{tab:dp}
  \begin{tabular}{@{}lcccccccc@{}}
    \toprule
    $\epsilon$ & 0.1 & 0.5 & 1 & 2 & 5 & 10 & +$\infty$ \\
    \midrule
    \sys & 0.3134 & 0.0773 & 0.0443 & 0.0210 & 0.0210 & 0.0122 & 0.0132 \\
    sync   & 0.3178 & 0.1253 & 0.0507 & 0.0468 & 0.0172 & 0.0110 & 0.0127 \\
    Honest Acc(\%) & 87.75 & 90.00 & 94.97 & 95.90 & 96.72 & 97.36 & 99.11\\
    \bottomrule
  \end{tabular}
  \vspace{-3mm}
\end{table}

\section{Potential detection of our attacks}
In this section, we explore potential detection mechanisms tailored to identify our proposed attacks. Our attack integrates label information into the gradients, rendering simple observations of gradient differences across various labels ineffective for detection. Specifically, our approach not only classifies the corresponding labels of embeddings but also distinguishes between true and fake embeddings generated by either the local encoder or the target model. Consequently, we hypothesize that the increased number of labels and the incorporation of a GAN-like structure will influence the distribution of gradient norms. To investigate this, we analyze the distribution of gradients' L2 norms across both tabular and image datasets.

\begin{figure}[htb] 
    \centering
    \begin{subfigure}[b]{0.24\textwidth}
        \centering
        \includegraphics[width=\linewidth]{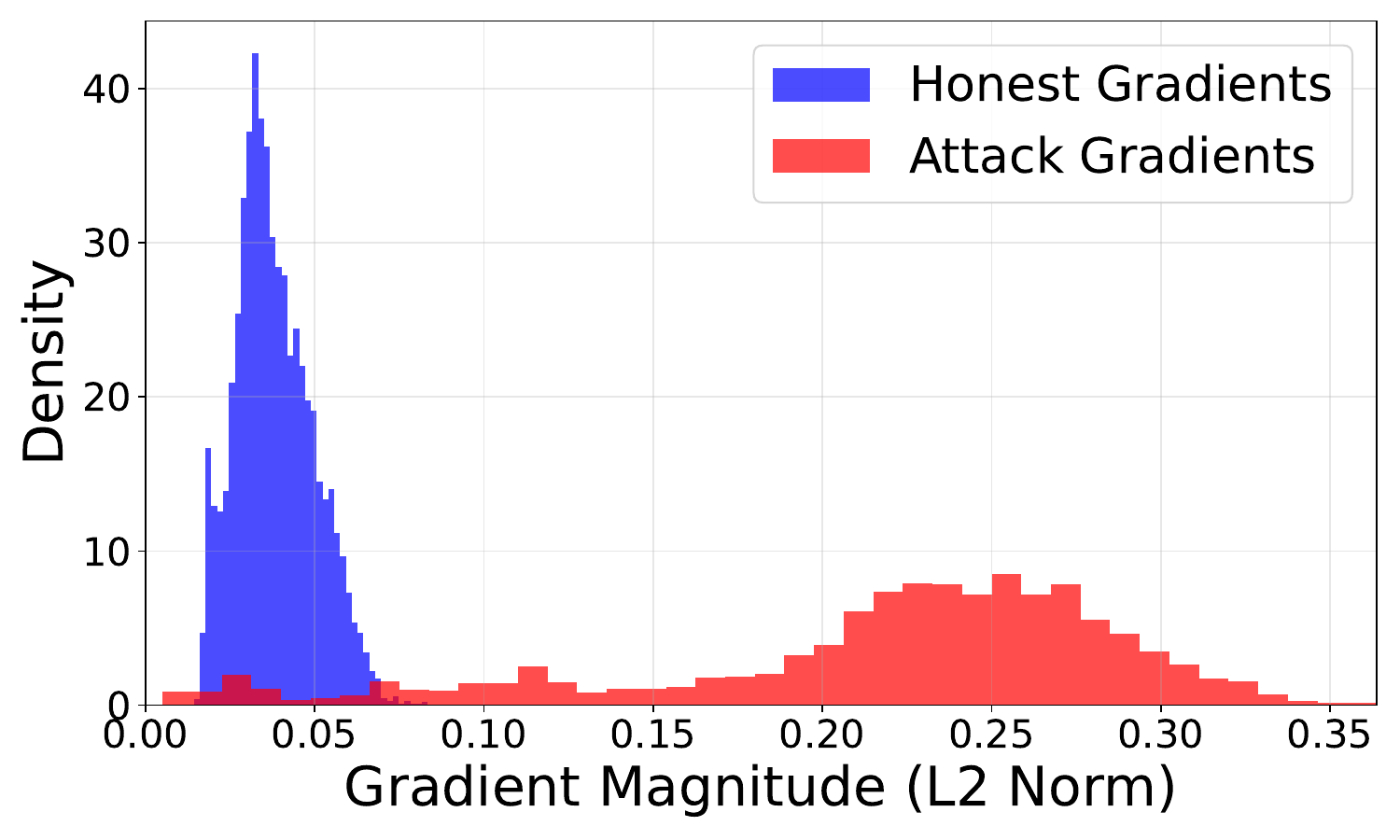} 
        \caption{Embedding gradients norm in Credit dataset.}
        \label{fig:figure1_credit}
    \end{subfigure}
    \hfill
    \begin{subfigure}[b]{0.24\textwidth}
        \centering
        \includegraphics[width=\linewidth]{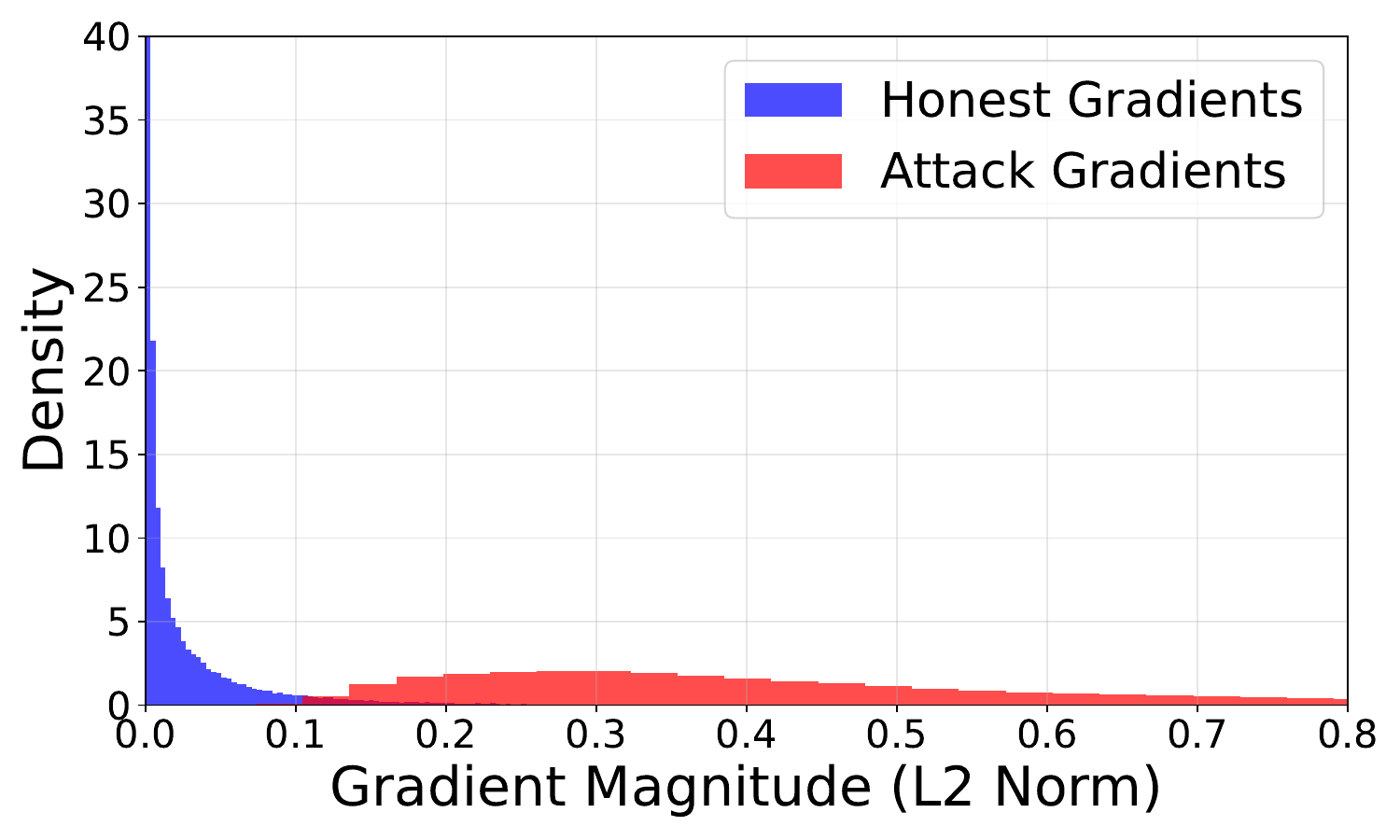} 
        \caption{Embedding gradients norm in IOT dataset.}
        \label{fig:figure2_iot}
    \end{subfigure}
    
    \vspace{10pt} 
    
    \begin{subfigure}[b]{0.24\textwidth}
        \centering
        \includegraphics[width=\linewidth]{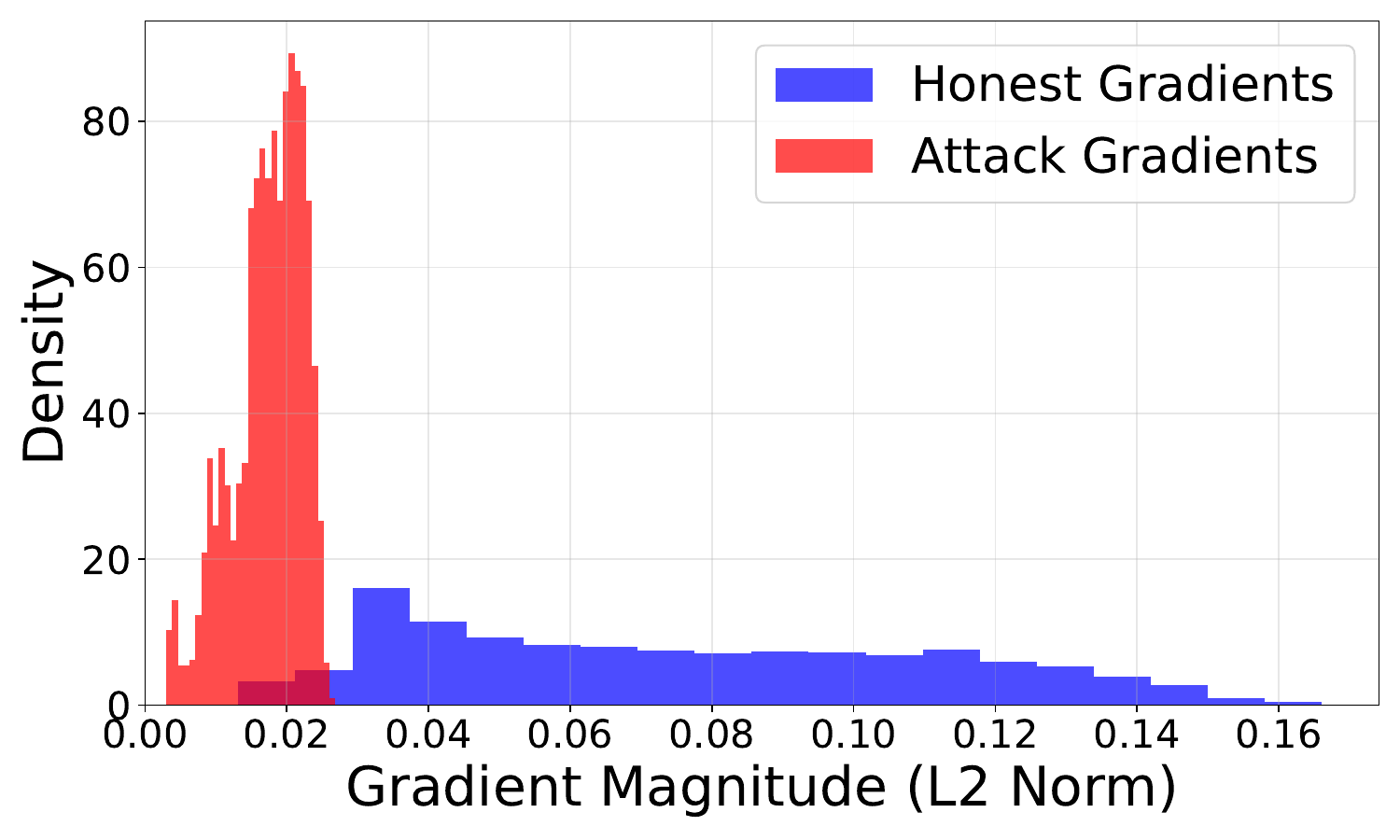} 
        \caption{Embedding gradients norm in CIFAR-10 dataset.}
        \label{fig:figure3_cifar}
    \end{subfigure}
    \hfill
    \begin{subfigure}[b]{0.24\textwidth}
        \centering
        \includegraphics[width=\linewidth]{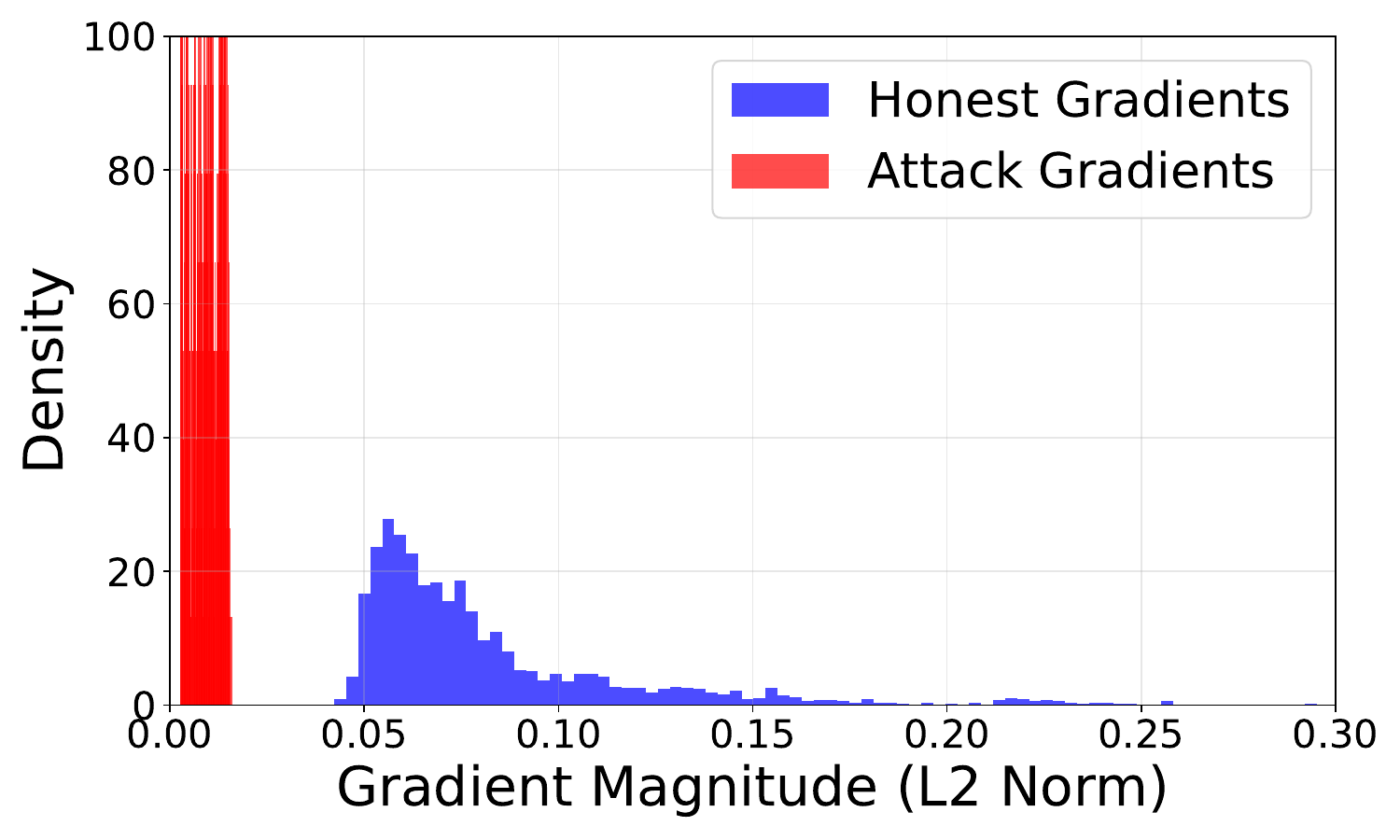} 
        \caption{Embedding gradients norm in Tiny imagenet dataset.}
        \label{fig:figure4_tiny}
    \end{subfigure}
    \caption{Distribution of Embedding Gradients' L2 Norms in Honest Training versus Our URVFL Attack.}
    \label{fig:designed_detection}
\end{figure}

Fig.\ref{fig:designed_detection} illustrates a clear distinction in the distribution of embedding gradients' L2 norms between honest training and our \sys attack. In tabular datasets, the gradient norms under attack are sparsely distributed with a higher average value, whereas, in honest training, they concentrate around lower values. Conversely, for image datasets, where the pattern reverses, honest training results in sparsely distributed gradients with higher average norms, while the \sys attack leads to concentrated gradients with lower average norms. This discrepancy is likely attributable to inherent differences in data characteristics and model architectures between image and tabular datasets.

Despite the varying gradient norm distributions between tabular and image datasets, a consistent and significant difference remains evident between honest training and the \sys attack across both data types. Leveraging this insight, a passive client can enhance detection capabilities by adopting a two-step strategy. Initially, the passive client collects publicly available comprehensive training data and conducts local training for several hundred iterations to establish a baseline distribution of embedding gradients' L2 norms under honest training conditions. This baseline serves as a reference point for normal gradient behavior.

During subsequent training sessions, the passive client continuously records the distribution of embedding gradients. By comparing the observed gradient distribution against the pre-established honest training distribution, the client can identify significant deviations indicative of an attack. If the gradient norms exhibit patterns that deviate substantially from the baseline, such as increased sparsity and higher average norms in tabular datasets or the opposite pattern in image datasets, the passive client can infer the presence of an attack. Upon detecting such discrepancies, the passive client can stop updating the local bottom model, thereby preventing further compromise and safeguarding the integrity and privacy of the training process.

%% file: conclusion.tex
\section{Conclusion}
We introduce \sys, a novel undetectable malicious data reconstruction attack for vertical federated learning (VFL). \sys innovatively incorporates training of a discriminator with auxiliary classifier (DAC) to 
integrate label information for malicious gradient generation, which not only makes the malicious training indistinguishable from honest training, but also substantially improves data reconstruction performance.
Through visualization and quantification, we demonstrated the effectiveness of DAC compared to traditional discriminators, showcasing its superior capability in embedding transfer. Through extensive experiments, we empirically demonstrated that \sys successfully evades SOTA detection methods and outperforms all existing data reconstruction attacks.
While traditional defense methods based on embedding/model perturbations are shown to effectively defend \sys, it is at the cost of serious degradation on the honest VFL task's performance. This calls for future research to develop effective defenses against malicious data reconstruction, with minimal impact on the honest task. 

%% file: appendix.tex
\section{URVFL\_sync algorithm}\label{appB}
We detail the steps of URVFL\_sync in Algorithm~\ref{alg3}.

\begin{algorithm}[h]

\KwInput{Auxiliary dataset $D_{aux}= \{\boldsymbol{x}_{i,a}, \boldsymbol{x}_{i,p}\}_{i=1}^{M_{aux}}$ and training dataset $D_{train} = \{\boldsymbol{x}_{j,a}, \boldsymbol{x}_{j,p}\}_{j=1}^{M}$.}

\KwOutput{Encoder $f_e(\cdot)$, DAC $D(\cdot)$, adversary's bottom model $f_a(\cdot)$, and target  model $f_p(\cdot)$.}

\textit{Adversary Procedure:}

\While{VFL training}
{

The adversary select a batch $\mathcal{B}_{aux}$ from $D_{aux}$;

Compute $\boldsymbol{h}_{i,a} = f_a(\boldsymbol{x}_{i,a}),i \in \mathcal{B}_{aux}$, and  $\boldsymbol{h}_{i,p} = f_e(\boldsymbol{x}_{i,p}), i \in \mathcal{B}_{aux}$;

Compute $\Tilde{\boldsymbol{x}}_{i, t} = f_d([\boldsymbol{h}_{i,a}\| \boldsymbol{h}_{i,p}]), i\in\mathcal{B}_{aux}$;

$\mathcal{L}_R \gets \frac{1}{|\mathcal{B}_{aux}|}\sum_{i\in \mathcal{B}_{aux} }MSE(\Tilde{\boldsymbol{x}}_{i, t}, \boldsymbol{x}_{i, t})$;

$\boldsymbol{\nabla}_e \gets$ Gradient($\mathcal{L}_R, f_e(\cdot)$);

$\boldsymbol{\nabla}_a \gets$ Gradient($\mathcal{L}_R, f_a(\cdot)$);

$\boldsymbol{\nabla}_d \gets$ Gradient($\mathcal{L}_R, f_d(\cdot)$);

$f_e(\cdot) \gets \text{Model\_update}(f_e(\cdot), \boldsymbol{\nabla}_e $);

$f_a(\cdot) \gets \text{Model\_update}(f_a(\cdot), \boldsymbol{\nabla}_a $);

$f_d(\cdot) \gets \text{Model\_update}(f_d(\cdot), \boldsymbol{\nabla}_d $);

Compute $\boldsymbol{h}_{i,p} = f_e(x_{i,p}), i \in \mathcal{B}_{aux}$;

All clients agree a batch data $\mathcal{B}_{train}$ from $D_{train}$;

Receive and record passive clients' embeddings $\boldsymbol{h}_{j,p},j \in \mathcal{B}_{train}$;

Compute the loss $\mathcal{L}_M = \frac{1}{|\mathcal{B}_{train}|}\sum_{j\in\mathcal{B}_{train}}CE(y_j^{+}, D(\boldsymbol{h}_{j,p}))$;

$\boldsymbol{\nabla}_{p}  \gets $Gradient($\mathcal{L}_M, \boldsymbol{h}_{j,p}, j \in \mathcal{B}_{train})$;

Send $\boldsymbol{\nabla}_{p}$ to the passive clients;

Compute the DAC loss $\mathcal{L}_D$ in (\ref{LD});

$\boldsymbol{\nabla}_{D}  \gets$  Gradient($\mathcal{L}_D, D(\cdot))$;

$D(\cdot) \gets \text{Model\_update}(D(\cdot), \boldsymbol{\nabla}_D $);
}

\textit{Passive Clients Procedure:}

\While{VFL training }{

All clients agree a batch data $\mathcal{B}_{train}$ from $D_{train}$;

Compute embeddings $\boldsymbol{h}_{j,p} = f_p(\boldsymbol{x}_{j,p}), j\in \mathcal{B}_{train}$ and send embeddings to the active client;

Receive gradient $\boldsymbol{\nabla}_p$;

$f_p(\cdot) \gets \text{Model\_update}(f_p(\cdot), \boldsymbol{\nabla}_p)$;
}

\caption{URVFL\_sync.}\label{alg3}
\end{algorithm}

\section{Datasets and model descriptions}\label{appendix_a}
We illustrate all datasets and models information used in the experiments in the following Table~\ref{tab:dataset_characteristics}, where MLP is composed of a linear and an activation, e.g., ReLU layer, CNN layer is composed a convolution process and a linear layer, and Residual block is constructed by 3 CNN layers and a skip connection.

\section{Visualization of reconstruction in Tiny imagenet}\label{tinyrecon}
In addition to visualizing the reconstruction performance on CIFAR-10, we compare the reconstruction performance of AGN, FSHA, and URVFL on the Tiny ImageNet dataset in Table~\ref{tab:visual_tiny}.

\begin{table*}[hb]
  \centering
  \caption{Dataset characteristics and models structure}
  \scriptsize
  \label{tab:dataset_characteristics}
  \small 
  \begin{tabular}{@{}lccccc@{}}
    \toprule
    Dataset & Credit & RT\_IOT2022 & MNIST & CIFAR-10 & Tiny imagenet \\ 
    \midrule
    \begin{tabular}[c]{@{}l@{}}Feature sizes of the adversary \\ and the passive clients $d_a, d_p$\end{tabular} & 11, 12 & 41, 42 & 392, 392 & 512, 512 & 2048, 2048 \\
    The size of training dataset & 19,091 & 89,920 & 54,546 & 45,455 & 72,728 \\
    The size of  the auxiliary dataset& 1,909 & 8,573 & 5,454 & 4,545 & 7,272\\
  The size of the test dataset & 9,000 & 24,624 & 10,000 & 10,000 & 20,000 \\
      $f_e(\cdot), f_a(\cdot),$ and  $f_p(\cdot)$ structure & MLP layer$\times$2 & MLP layer$\times$2 & CNN layer$\times$2 & Residual block$\times$3&Residual block$\times$3\\
    $f_d(\cdot)$ structure & MLP layer$\times$3& MLP layer$\times$3 & CNN layer$\times$3 & CNN layer$\times$3 &CNN layer$\times$4\\
    $D(\cdot)$ structure & MLP layer$\times$3& MLP layer$\times$3& CNN layer$\times$2 & CNN layer $\times$3&CNN layer$\times$3\\
    Learning rate&1e-4& 1e-4&1e-4&1e-3&1e-4\\
    \sys pretraing epochs&30&40&10&20&30\\
    \bottomrule
  \end{tabular}
\end{table*}

\begin{table*}[ht]
\centering
\caption{Visualization of reconstructed features on the Tiny imagenet dataset with and without the detection SG. Except for the original image, the right side of each image is the reconstructed image. The last column is the average reconstruction error. }
\includegraphics[width=1\linewidth]{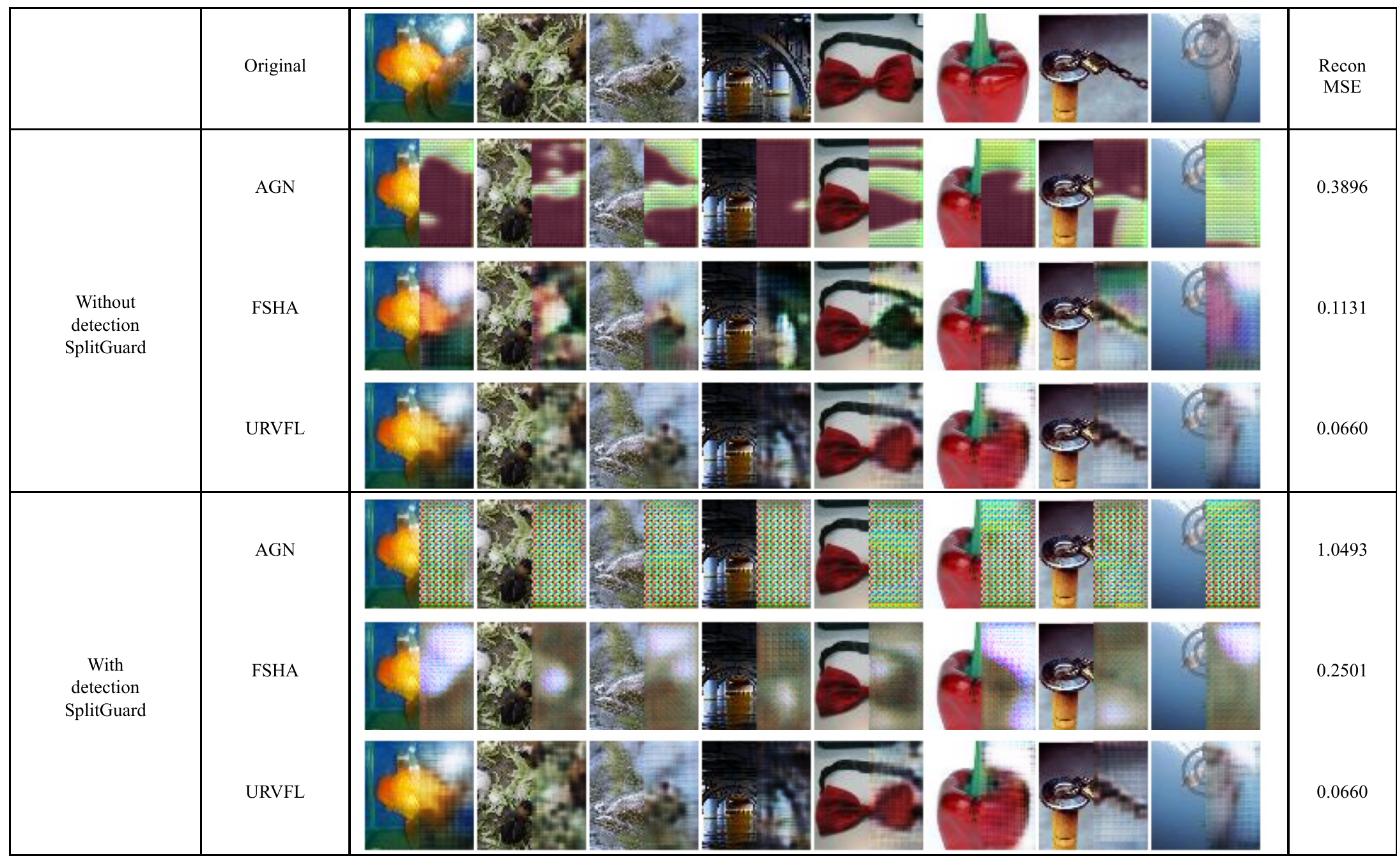}
\label{tab:visual_tiny}
\end{table*}